\documentclass[letterpaper]{article} %
\usepackage{aaai2027}  %
\usepackage[hyphens]{url}  %
\usepackage{graphicx} %
\usepackage{natbib}  %
\usepackage{caption} %
\usepackage{algorithm}
\usepackage{algorithmic}

\usepackage{newfloat}
\usepackage{listings}
\DeclareCaptionStyle{ruled}{labelfont=normalfont,labelsep=colon,strut=off} %
\floatstyle{ruled}
\newfloat{listing}{tb}{lst}{}
\floatname{listing}{Listing}

\usepackage{booktabs}
\usepackage{graphicx}
\usepackage{multirow}
\usepackage{enumitem}
\usepackage{capt-of}
\usepackage{tabularx}
\usepackage{amsthm}
\usepackage{amsmath}
\usepackage{amssymb}
\usepackage{subcaption}

\newcommand{\modelname}{\texttt{CADonatello}}

\title{\raisebox{-0.21\height}{\includegraphics[width=0.045\textwidth]{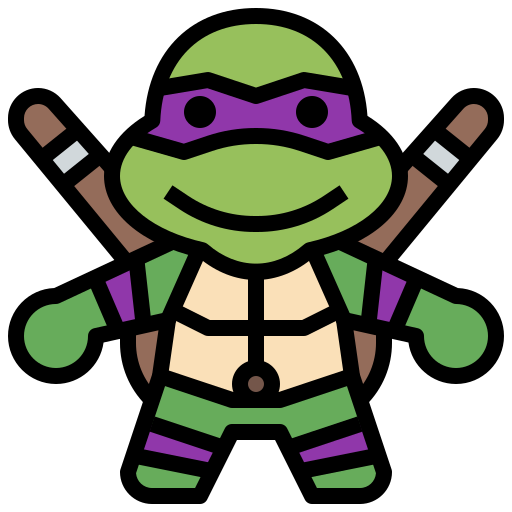}}~\texttt{CADonatello}: Towards Real-World Multi-Part 3D CAD Reconstruction}
\author{
    Ahmet Serdar Karadeniz\textsuperscript{\rm 1}, Dimitrios Mallis\textsuperscript{\rm 1}, Parsa Akbari Chianeh\textsuperscript{\rm 1}, Kseniya Cherenkova\textsuperscript{\rm 2},\\ Anis Kacem\textsuperscript{\rm 1}, Djamila Aouada\textsuperscript{\rm 1}
}
\affiliations{
    \textsuperscript{\rm 1}SnT, University of Luxembourg, \textsuperscript{\rm 2}Artec 3D\\

}

\begin{document}

\maketitle

\begin{abstract}
Existing methods for CAD reconstruction from 3D scans achieve strong performance on standard benchmarks, yet their effectiveness often degrades in real-world scenarios where objects consist of multiple parts and scan artifacts such as noise. A key reason is that current CAD reconstruction datasets predominantly contain single-part models and clean geometric inputs, limiting the ability of learned models to generalize to realistic settings. In this work, we introduce~\modelname, a part-aware framework for reconstructing parametric CAD models from real-world 3D scans. Our approach first decomposes the input geometry into structurally meaningful components using a foundation-scale 3D part segmentation model, and then reconstructs each component independently through a cross-section–driven pipeline that performs sketch parameterization followed by an extrusion. To improve the robustness of sketch parameterization to scan-derived cross-sections, we further introduce \texttt{CC3D Cross-Sections}, a dataset of paired rasterized cross-section images from realistic scans and their corresponding parametric sketches derived from aligned CAD models. This dataset enables effective domain adaptation from clean CAD sketch images to scan-derived cross-section images. Extensive experiments demonstrate that the proposed approach improves generalization to noisy scans and enables reliable reconstruction of complex multi-part objects across multiple real-world CAD reverse engineering benchmarks.

\end{abstract}

\section{Introduction}

Computer-Aided Design (CAD) is a fundamental step in manufacturing everyday objects, where components are typically designed in CAD software before production. In practice, however, the original CAD models are not always available once an object has been manufactured due to legacy processes, data loss, proprietary restrictions, or undocumented modifications~\cite{beniere2013comprehensive}. With the widespread availability of 3D scanning technologies, physical objects can be digitized as point clouds or meshes~\cite{cherenkova2023sepicnet}. However, these geometric representations do not directly provide editable, parametric CAD models. Recovering CAD models from scan data, \textit{i.e.}, CAD reverse engineering, is therefore essential for redesign, modification, and remanufacturing within standard CAD workflows~\cite{khan2024cad}.

\begin{figure}[t!]
    \centering
    \includegraphics[width=\linewidth]{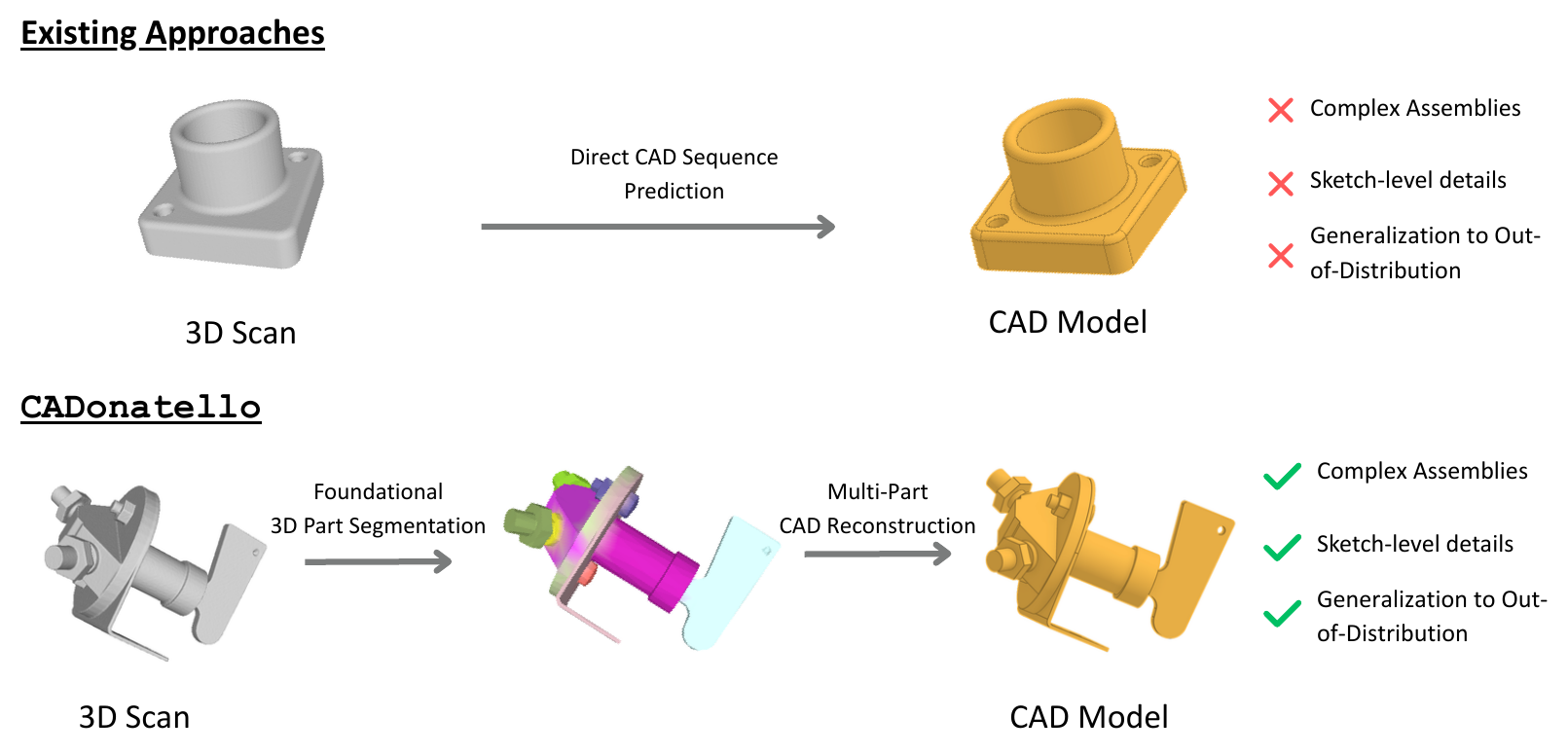}
    \caption{\modelname~is a part-aware CAD reverse engineering method that decomposes 3D scans using a foundation model for 3D part segmentation and reconstructs editable CAD models through cross-section–driven parametric part reconstruction.}
    \label{fig:teaser}
    \vspace{-0.5cm}
\end{figure}

Current reverse engineering pipelines rely on specialized software that imports 3D scans as geometric references inside a CAD environment. Designers manually reconstruct the model using parametric sketching and feature-based operations such as extrusion and revolution~\cite{mallis2023sharp}. While effective, this process is inherently time-consuming and labor-intensive, particularly for geometrically complex or multi-part assemblies. Automating CAD reconstruction from 3D scans has thus emerged as a key research challenge~\cite{wu2021deepcad, uy2022point2cyl, ren2022extrudenet, ma2024draw, khan2024cad, rukhovich2025cad, karadenizMiCADangelo, kolodiazhnyi2025cadrille}. Formally, the goal is: given a 3D scan represented as a point cloud or mesh, predict the parametric CAD construction sequence that reproduces the observed geometry. Such a sequence typically comprises 2D sketches defined by closed profiles of parametric primitives (e.g., lines, arcs), followed by 3D solid modeling operations such as extrusion and revolution.

Recent CAD datasets, such as DeepCAD and Fusion 360 reconstruction~\cite{wu2021deepcad, willis2020fusion}, have accelerated research on this problem by providing access to parametric construction sequences, typically structured as sketch-and-extrude programs. These resources enabled learning-based approaches that map 3D scans to CAD operation sequences either through geometry-grounded extrusion reasoning~\cite{uy2022point2cyl, karadenizMiCADangelo} or via end-to-end program inference~\cite{dupont2024transcad, khan2024cad, rukhovich2025cad, kolodiazhnyi2025cadrille}. Although these approaches achieved strong performance on benchmark datasets, their generalization capability remains inherently constrained by the distributional scope of their training data. First, existing datasets predominantly consist of single-part CAD models with relatively simple feature hierarchies. Consequently, learned models are biased toward such simplified design patterns and struggle to scale to complex, multi-part assemblies or industrial-grade geometries that deviate from the training distribution. Second, these datasets typically rely on synthetic, clean geometric inputs and do not provide realistic 3D scans containing acquisition artifacts such as noise, occlusions, and incomplete surfaces.

In this work, we argue that enabling real-world CAD reverse engineering of complex, multi-part assemblies from noisy 3D scans requires moving beyond learning a direct mapping from input geometry to CAD sequences using the aforementioned limited datasets. As shown in Figure~\ref{fig:teaser}, we adopt a structural decomposition paradigm that mirrors conventional design practice: we first segment the input geometry into semantically meaningful parts, reconstruct each part parametrically, and subsequently assemble the recovered components. To improve generalization beyond current CAD sequence datasets, we rely on models trained on large-scale data for both geometric decomposition and parametric reconstruction. Specifically, we leverage a foundation-scale 3D segmentation model~\cite{ma2025p3} to identify part-level structure in the input scan. For parametric reconstruction, we build on the cross-section paradigm of~\cite{karadenizMiCADangelo} and adapt it to part-level reconstruction through orientation-aware cross-section plane detection, image-level cross-section parameterization, and adaptive extrusion optimization. The cross-section parameterizer is pretrained on SketchGraphs~\cite{seff2020sketchgraphs} and further fine-tuned on paired scan-derived cross-sections and BRep sketches from CC3D~\cite{mallis2023sharp,cherenkova2020pvdeconv}, improving robustness to realistic scanning artifacts. The main contributions of this work are as follows:

\begin{enumerate}%
    \item \modelname{}, a novel CAD reverse engineering framework that formulates real-world reverse engineering as a structured, part-aware reconstruction problem, combining foundation-scale 3D segmentation with cross-section–driven parametric part modeling within a unified modular pipeline.

    \item \texttt{CC3D Cross-Sections}, a newly curated dataset of $44$k cross-sections comprising paired rasterized sketches extracted from realistic CC3D scans and their corresponding parametric representations derived from aligned BReps. The dataset explicitly captures scan-induced artifacts and geometric distortions at the sketch level, enabling the learning and evaluation of robust cross-section parameterization under realistic noise. The dataset will be publicly released.

    \item Extensive experimental validation in realistic CAD reverse engineering settings, demonstrating state-of-the-art performance on noisy CC3D scans and, for the first time, on complex multi-part assemblies using the Fusion 360 Assembly dataset.
\end{enumerate}

\section{Related Work}
\label{sec:related_work}

\noindent\textbf{3D CAD Reverse Engineering.} Early approaches to CAD reconstruction relied on Constructive Solid Geometry (CSG) representations~\cite{du2018inversecsg, friedrich2019optimizing, kania2020ucsg, geometry2022neural, yu2022capri, yu2023d, ren2021csg} based on Boolean operations over primitives, while later works focused on Boundary Representation (BRep) modeling~\cite{lambourne2021brepnet, willis2020fusion, jayaraman2021uv, xu2021inferring} to recover precise surfaces and topology. More recent methods learn sketch–extrude sequences, either through geometry-grounded approaches~\cite{uy2022point2cyl, ren2022extrudenet, li2023secad, karadenizMiCADangelo} or language-based inference~\cite{khan2024cad, rukhovich2025cad, ma2024draw}. Although effective on existing CAD benchmarks~\cite{wu2021deepcad, willis2020fusion}, these approaches typically reconstruct the entire object directly, which makes them fragile when applied to real-world scans containing multiple components. In contrast, \modelname{} operates at the level of structural parts, enabling parametric reconstruction of multi-part objects.

\noindent \textbf{3D Part Segmentation.} Segmenting 3D objects into meaningful components is fundamental for geometric understanding. Early approaches~\cite{qi2017pointnet++, wu2024point, hanocka2019meshcnn} focused on supervised learning with part-level annotations~\cite{mo2019partnet,li2022_3dcompat,slim2023_3dcompatplus,li20223d} on point clouds or meshes to predict predefined semantic parts, but generalize poorly to unseen categories. Later work leverages large-scale 2D vision models such as SAM~\cite{kirillov2023segment} to bootstrap 3D segmentation~\cite{tang2024segment,yang2023sam3d,xu2025sampro3d,yang2024sampart3d,liu2025partfield}. These approaches inherit the limitations of 2D priors, often leading to viewpoint ambiguities and inconsistent 3D boundaries~\cite{ma2025p3}. Recent foundation-scale native 3D models~\cite{ma2025p3, zhu2025partsam} learn part segmentation directly in 3D, enabling category-agnostic decompositions and generalization to unseen objects. Despite strong results in part segmentation, their potential for CAD reverse engineering remains largely unexplored. \modelname{} leverages such models to obtain part-level decompositions of scans and enable part-aware reconstruction of multi-component CAD models.

\noindent\textbf{Parametric CAD Sketches.} CAD sketches play an important role in 3D CAD reverse engineering, as they enable editability and support controlled modifications of CAD models. The emergence of large-scale sketch datasets~\cite{seff2020sketchgraphs, ganin2021computer} has significantly advanced CAD sketch research, enabling progress in generative sketch modeling~\cite{ganin2021computer,seff2021vitruvion,para2021sketchgen,willis2021engineering,chereddy2025sketchdnn}, inferring design intent~\cite{casey2025aligning,yang2022discovering}, and image-based sketch parameterization~\cite{ganin2021computer,seff2021vitruvion,para2021sketchgen,karadeniz2025picasso,karadeniz2024davinci,wu2024cadvlm,wang2024parametric}. Image-conditioned approaches~\cite{karadeniz2024davinci,karadeniz2025picasso} recover parametric sketches from 2D image inputs by predicting geometric primitives and constraints. More closely related to CAD reverse engineering,~\cite{karadenizMiCADangelo} infers parametric sketches from scan-derived cross-section images. However, this method relies on synthetic noise augmentations of clean sketch images for training, which may not fully reflect the geometric irregularities present in real cross-sections. In contrast, our sketch parameterization model is fine-tuned on the introduced \texttt{CC3D Cross-Section} dataset with real cross-sections paired with parametric sketches, improving robustness to real-world scans.

\section{Problem Statement}
\label{sec:prblem}

Given a 3D mesh \( \mathbf{M} = (\mathcal{V}, \mathcal{F}) \), where \( \mathcal{V} \subset \mathbb{R}^3 \) denotes the set of vertices and \( \mathcal{F} \subset \mathcal{V}^3 \) the set of triangular faces, the objective is to recover a structured CAD representation defined as a sequence of $n_s$ sketch-extrude steps that reconstruct the target geometry $
\mathbf{C} = \{(\mathcal{K}_j,\mathbf{e}_j)\}_{j=1}^{n_s}$, with: 
\begin{enumerate}
    \item[(1)] each sketch $\mathcal{K}_j$ is defined on a sketch plane $\pi_j$, specified by an origin $\mathbf{o}_j \in \mathbb{R}^3$ and a normal vector $\mathbf{n}_j \in \mathbb{R}^3$. The sketch $\mathcal{K}_j = (\mathcal{P}_j,\mathcal{C}_j)$ encodes:
    \begin{enumerate}
        \item[(i)] the set of geometric primitives $\mathcal{P}_j = \{\mathbf{p}_i\}_{i=1}^{n_p}$ consisting of $n_p$ 2D parametric entities uniquely defined by geometric parameters as follows. A \textit{line} primitive is defined by its start point $(x_s, y_s) \in \mathbb{R}^2$ and end point $(x_e, y_e) \in \mathbb{R}^2$. A \textit{circle} primitive is defined by its center $(x_c, y_c) \in \mathbb{R}^2$ and radius $r \in \mathbb{R}$. An \textit{arc} primitive is defined by its start point $(x_s, y_s) \in \mathbb{R}^2$, mid-point $(x_m, y_m) \in \mathbb{R}^2$, and end point $(x_e, y_e) \in \mathbb{R}^2$, where the three points are non-collinear and lie on a common circle. 
        \item[(ii)] and their associated constraints $\mathcal{C}_j = \{\mathbf{c}_i\}_{i=1}^{n_c}$ encoding geometric relationships between primitives, where each constraint $\mathbf{c} = (\mathbf{p}_i,\mathbf{p}_k,c_t)$ specifies a type $c_t$ such as \textit{coincident, concentric, equal, horizontal, midpoint, normal, parallel, perpendicular, tangent, or vertical}.
    \end{enumerate}

    \item[(2)] each extrusion $\mathbf{e}_j = (\pi_j, t_j, \mathbf{v}_j, h_j)$ extends the sketch $\mathcal{K}_j$ along a direction vector $\mathbf{v}_j \in \mathbb{R}^3$ for an extrusion height $h_j \in \mathbb{R}$, where the extrusion type $t_j$ specifies whether material is added (\textit{new}) or removed (\textit{cut}).

\end{enumerate}

Therefore, the problem consists of recovering the sequence 
\(
\mathbf{C} = \{(\mathcal{K}_j,\mathbf{e}_j)\}
\),
including the sketch planes $\{\pi_j\}$, constrained sketches $\{\mathcal{K}_j\}$, and extrusions $\{\mathbf{e}_j\}$, whose composition reconstructs the input mesh $\mathbf{M}$.

\section{Proposed Method}
\label{sec:method}

\begin{figure*}[t]
    \centering
    \includegraphics[width=0.8\linewidth
    ]{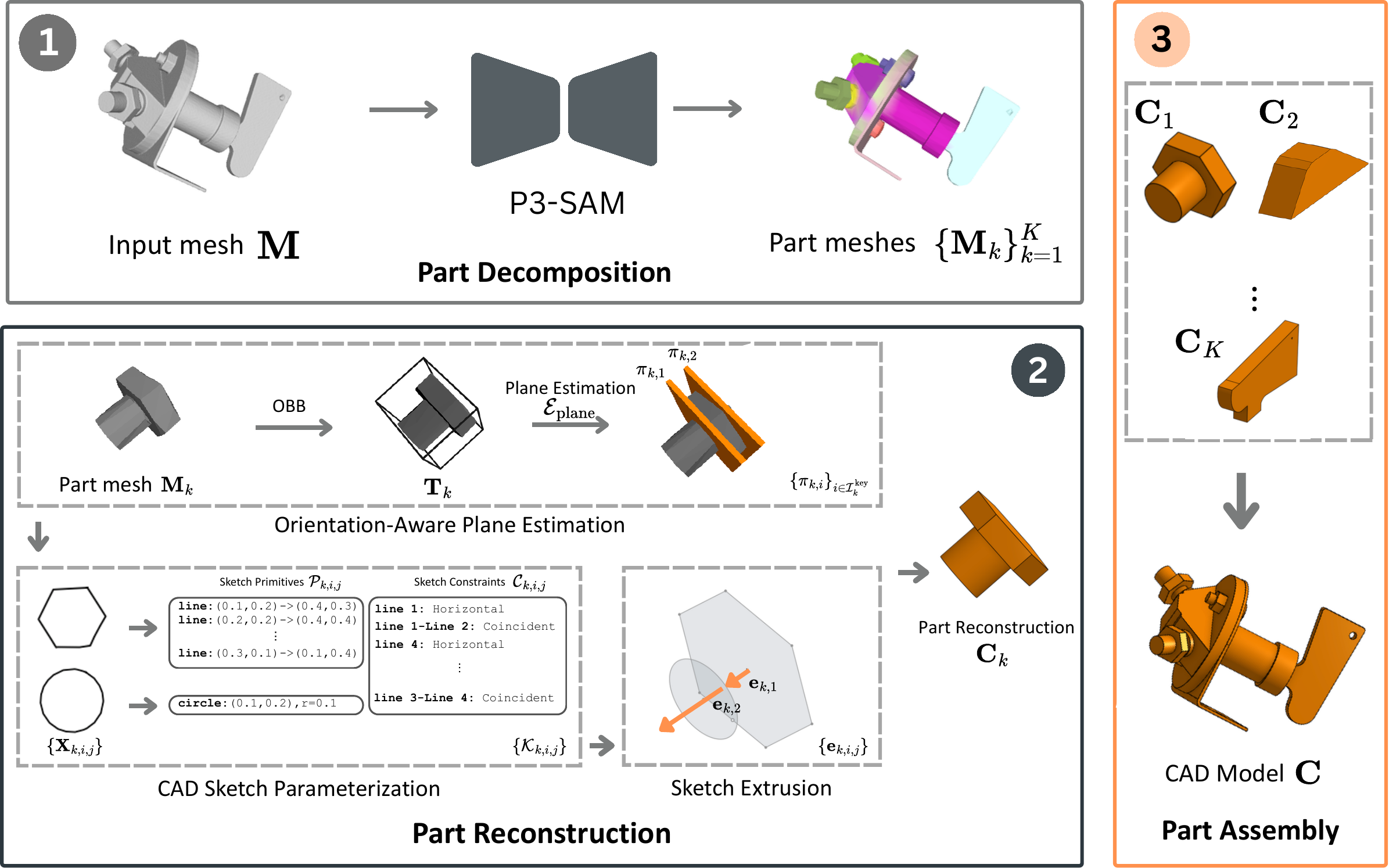}
    \caption{Overview of CADonatello. The input mesh is decomposed
    into parts, each part is reconstructed through cross-section
    parameterization and extrusion inference, and the recovered
    parts are assembled into the final CAD model.}
    \label{fig:overview}
    \vspace{-0.6cm}
\end{figure*}

\modelname{} tackles the problem with a focus on real-world scenarios where input meshes $\mathbf{M}$ exhibit complex multi-part assemblies and scanning artifacts. Rather than directly learning the mapping between the input mesh $\mathbf{M}$ and the corresponding CAD sequence $\mathbf{C} = {(\mathcal{K}_j,\mathbf{e}_j)}_{j=1}^{n_s}$ from existing limited datasets, \modelname{} first decomposes the input mesh $\mathbf{M}$ into a set of $K$ structurally meaningful parts $\{\mathbf{M}_k\}_{k=1}^{K}$. To ensure generalization, \modelname{} leverages a foundational model to perform this decomposition. Each part $\mathbf{M}_k = (\mathcal{V}_k,\mathcal{F}_k)$ is then reconstructed independently by estimating a part-level CAD sequence $\mathbf{C}_k$ that reconstructs $\mathbf{M}_k$. For this stage, part reconstruction follows a cross-section-based approach similar to~\cite{karadenizMiCADangelo}, where constrained sketch parameterization is learned at the raster sketch level using a large-scale dataset, and extrusions are inferred directly from the input geometry. To improve robustness to noisy scanning artifacts, the sketch parameterization module is further fine-tuned on a curated dataset of cross-sections derived from realistic 3D scans. The final CAD model $\mathbf{C}$ is obtained by assembling the reconstructed part-level sequences $\{\mathbf{C}_k\}_{k=1}^{K}$ according to their original spatial configuration. An overview of the proposed method is illustrated in Figure~\ref{fig:overview}, and the details of the different steps are provided in the following sections.

\vspace{-0.2cm}
\subsection{Part Decomposition}

To obtain the set of structurally meaningful part meshes  $\{\mathbf{M}_k\}_{k=1}^{K}$ from the input mesh $\mathbf{M}$, we leverage the foundation-scale 3D part segmentation model P3-SAM~\cite{ma2025p3}, which operates directly on 3D geometry and automatically produces candidate part masks. Unlike prior CAD reconstruction approaches that operate on complete objects or decompose them according to the underlying CAD operations, we use category-agnostic part masks to define structurally meaningful units of parametric reconstruction. P3SAM follows the paradigm of the Segment Anything Model (SAM)~\cite{kirillov2023segment} for 2D images, but is designed for 3D data. It is trained to segment 3D models into parts using a large-scale dataset containing more than $3$ million examples. Given an input mesh, P3SAM performs automatic part segmentation by sampling a uniform set of prompt points on the surface and predicting segmentation masks associated with the corresponding parts. Formally, let $\mathbf{Q} \in \mathbb{R}^{N_q \times 3}$ denote a set of $N_q$ prompt points sampled from the surface of $\mathbf{M}$. P3SAM defines the mapping

\begin{equation}
\Phi_{\text{mask}}:
\mathbb{R}^{N_q \times 3} 
\rightarrow \mathbf{\{0,1\}}^{|\mathcal{F}| \times K},
\;
\mathbf{B}^{\text{face}}_{\text{mask}} = \Phi_{\text{mask}}(\mathbf{Q}) \ .
\end{equation}

\noindent that maps the input prompt points $\mathbf{Q}$ to a binary mask matrix $\mathbf{B}^{\text{face}}_{\text{mask}}$ denoting per-face part segmentation assignment. These masks  induce part submeshes 
$\mathbf{M}_k = (\mathcal{V}_k,\mathcal{F}_k)$ defined as
\begin{equation}
\begin{aligned}
\mathcal{F}_k &= \{ f \in \mathcal{F} \mid 
\mathbf{B}^{\text{face}}_{\text{mask}}(f,k) = 1 \} \ ,\\
\mathcal{V}_k &= \{ \mathbf{v} \in \mathcal{V} \mid 
\exists f \in \mathcal{F}_k \ \text{such that} \ \mathbf{v} \in f \} \ .
\end{aligned}
\end{equation}
The resulting set $\{\mathbf{M}_k\}_{k=1}^{K}$ forms a geometrically 
coherent decomposition of the input mesh, enabling independent 
parametric reconstruction of each part.

\vspace{-0.2cm}
\subsection{Part Reconstruction}
\label{sec:partrec}

Given a segmented part mesh $\mathbf{M}_k$, the goal of part reconstruction is to estimate a part-level CAD sequence $\mathbf{C}_k = {(\mathcal{K}_{k,j},\mathbf{e}_{k,j})}_{j=1}^{n_k}$ that reconstructs $\mathbf{M}_k$, where $n_k$ denotes the number of sketch--extrude operations associated with the part $\mathbf{M}_k$. To achieve this, we build on the cross-section paradigm of~\cite{karadenizMiCADangelo}, while reformulating it for structurally decomposed parts through orientation-aware key-plane estimation and adaptive single- or multi-extrusion reconstruction. Specifically, the reconstruction is performed by (1) identifying key slicing planes that produce cross-sectional sketch images; (2) parameterizing the constrained cross-sections; and (3) inferring the extrusion operations corresponding to the different cross-sections.

\vspace{-0.1cm}
\subsubsection{Orientation-aware Key Slicing Plane Estimations.}
A segmented part mesh $\mathbf{M}_k = (\mathcal{V}_k,\mathcal{F}_k)$ has its own orientation and location relatively to the overall mesh $\mathbf{M}$.  The aim of key plane estimation is to obtain a set of key slicing planes $\{\pi_{k,i}\}_{i=1}^{n_{\text{key}}}$ in the original coordinate system of the part, allowing the handling of arbitrary orientations. To do so, we first canonicalize the part mesh $\mathbf{M}_k$ by computing its oriented bounding box (OBB), which defines a local coordinate system aligned with the principal directions of the part. Let $\mathbf{T}_k = (\mathbf{R}_k,\mathbf{t}_k) \in \mathcal{SE}(3)$ denote the Euclidean rigid transformation associated with the OBB, where $\mathbf{R}_k \in \mathcal{SO}(3)$ is the rotation matrix formed by the OBB principal axes and $\mathbf{t}_k \in \mathbb{R}^3$ is the OBB center. This transformation allows mapping a point from the canonical coordinate system to the OBB frame and the canonicalized version of the part mesh $\tilde{\mathbf{M}}_k = (\tilde{\mathcal{V}}_k,\mathcal{F}_k)$ is obtained by applying the inverse transformation 

\begin{equation}
\tilde{\mathbf{x}} = \mathbf{T}_k^{-1}(\mathbf{x}), 
\quad \forall \mathbf{x} \in \mathcal{V}_k, \ \tilde{\mathbf{x}} \in \tilde{\mathcal{V}}_k \ .
\end{equation}
To obtain key slicing planes from the canonicalized mesh $\tilde{\mathbf{M}}_k$, 
we define a plane estimator $\mathcal{E}_{\text{plane}}$ as a mapping that assigns 
to each $\tilde{\mathbf{M}}_k$ a set of $n_{\text{key}}$ slicing planes:
$
\mathcal{E}_{\text{plane}}(\tilde{\mathbf{M}}_k)
=
\{\pi_{k,i}\}_{i=1}^{n_{\text{key}}},
\quad
\pi_{k,i} = (\mathbf{o}_{k,i}, \mathbf{n}_{k,i}) 
\in \mathbb{R}^3 \times \mathbb{R}^3.
$ Importantly, if a part can be reconstructed in a single step, \textit{i.e.}, using a single 
sketch–extrude operation, only one key slicing plane should be estimated by 
$\mathcal{E}_{\text{plane}}$, avoiding the need to predict multiple planes and 
infer the corresponding sketches and extrusions. For this reason, we consider two slicing plane estimation strategies: one that identifies a single key slicing plane from the canonical planes ($\mathcal{E}_{\text{plane}}^{\text{single}}$), and another that predicts multiple key slicing planes ($\mathcal{E}_{\text{plane}}^{\text{multi}})$. The choice between these strategies is determined after performing the corresponding extrusions and comparing the resulting solid with the input geometry (see the adaptive extrusions subsection).

\vspace{-0.1cm}
\paragraph{Canonical single plane candidates  ($\mathcal{E}_{\text{plane}}^{\text{single}}$).} If a part $\tilde{\mathbf{M}}_k$ can be reconstructed with a single sketch-extrude operation, only three key slicing plane candidates are considered. In particular, the three plane candidates correspond to the planes passing by the origin $\mathbf{0}=(0,0,0)$ and having as normal one of the three canonical axes 
$\mathbf{a}_1=(1,0,0)$, 
$\mathbf{a}_2=(0,1,0)$, or 
$\mathbf{a}_3=(0,0,1)$. The corresponding canonical planes are
$
\tilde{\pi}_{k,r} = (\mathbf{0}, \mathbf{a}_r)$ where $r \in \{1,2,3\}.$
Mapping these planes back to the original coordinate system yields
 $\mathcal{E}_{\text{plane}}^{\text{single}}(\tilde{\mathbf{M}}_k)
=
\{\pi_{k,r}\}_{r=1}^{3},
\pi_{k,r}
=
\big(
\mathbf{t}_k,
\mathbf{R}_k \mathbf{a}_r
\big).$

\vspace{-0.1cm}
\paragraph{Learning-based multi-plane estimator ($\mathcal{E}_{\text{plane}}^{\text{multi}}$).}
If a part $\tilde{\mathbf{M}}_k$ cannot be accurately reconstructed using a single sketch--extrude operation, a learning-based plane detector is employed to identify key slicing planes in the canonical coordinate frame. This detector follows the strategy proposed in~\cite{karadenizMiCADangelo}. A set of $N$ candidate cross-section slicing planes $\{\tilde{\mathcal{S}}_{k,i}\}_{i=1}^{N}$ is uniformly sampled along the canonical $\mathbf{a}_1$, $\mathbf{a}_2$, and $\mathbf{a}_3$ axes. A transformer-based network $f_{\text{key}}$, trained on the DeepCAD dataset~\cite{wu2021deepcad}, then assigns key-plane probabilities

\begin{equation}
\hat{y}_{k,i} = f_{\text{key}}(\tilde{\mathcal{S}}_{k,i}),
\qquad
\mathcal{I}_k^{\text{key}} = \{ i \mid \hat{y}_{k,i} \ge \tau \} \ ,
\end{equation}
where $\tau \in [0,1]$ denotes a probability threshold. For more details about $f_{\text{key}}$ readers are referred to~\cite{karadenizMiCADangelo}. For each $i \in \mathcal{I}_k^{\text{key}}$, the corresponding slice induces a key plane $
\tilde{\pi}_{k,i} = (\tilde{\mathbf{c}}_{k,i}, \tilde{\mathbf{n}}_{k,i})
\in \mathbb{R}^3 \times \mathbb{R}^3 .
$ Mapping these planes back to the original coordinate system yields $
\mathcal{E}_{\text{plane}}^{\text{multi}}(\tilde{\mathbf{M}}_k)
=
\{\pi_{k,i}\}_{i \in \mathcal{I}_k^{\text{key}}},
\pi_{k,i}
=
\big(
\mathbf{R}_k \tilde{\mathbf{c}}_{k,i} + \mathbf{t}_k,
\mathbf{R}_k \tilde{\mathbf{n}}_{k,i}
\big).$

\vspace{-0.1cm}
\subsubsection{Scan Cross-section Parameterization.} 
Given a part mesh $\mathbf{M}_k$ and a selected key slicing plane $\pi_{k,i}$, the mesh is intersected with the plane to obtain a 2D cross-sectional profile. The goal is to parameterize this profile using parametric primitives together with their associated geometric constraints. To achieve this, we define a part-level rasterization operator
\begin{equation}
\mathcal{R}(\mathbf{M}_k,\pi_{k,i})
=
\{\mathbf{X}_{k,i,j}\}_{j=1}^{n_l^{k,i}},
\;
\mathbf{X}_{k,i,j} \in \mathbf{\{0,1\}}^{H \times W} \ ,
\end{equation}
which generates binary loop images $\mathbf{X}_{k,i,j}$ of size $H \times W$ from the intersections 
$\mathbf{M}_k \cap \pi_{k,i}$ through loop extraction, 2D projection, and rasterization. Here, $n_l^{k,i}$ denotes the number of loop images obtained from the key plane $\pi_{k,i}$. The constrained sketch parameterization network follows the architecture introduced in~\cite{karadeniz2024davinci}. Each loop image $\mathbf{X}_{k,i,j}$ is processed by a transformer-based encoder $f_{\text{encoder}}$ followed by two prediction heads that estimate primitives and constraints:
\begin{equation}
\begin{aligned}
\mathcal{P}_{k,i,j}
&= f_{\mathrm{prim}}\!\left(f_{\mathrm{encoder}}(\mathbf{X}_{k,i,j})\right),\\
\mathcal{C}_{k,i,j}
&= f_{\mathrm{ctr}}\!\left(f_{\mathrm{encoder}}(\mathbf{X}_{k,i,j})\right).
\end{aligned}
\end{equation}
As in~\cite{karadeniz2024davinci}, this constrained sketch parameterization network is pretrained on clean rasterized CAD sketch images from the large-scale SketchGraphs~\cite{seff2020sketchgraphs} dataset (1.5 million sketches). Since this pretraining is performed on clean rasterized sketches, a domain gap arises when applying the model to scan-derived cross-sections. To mitigate this issue, we construct a dedicated cross-section dataset derived from realistic 3D scans and fine-tune the parameterization model on it. Note that at inference, the constrained sketch parameterization is performed for each loop $\mathbf{X}_{k,i,j}$ of each key slicing plane $\pi_{k,i}$ of each part $\mathbf{M}_k$.

\vspace{-0.1cm}
\subsubsection{Adaptive Extrusion.}

Given a part mesh $\mathbf{M}_k$ and candidate key slicing planes for this part, either canonical planes $\{\pi_{k,r}\}_{r=1}^{3}$ or learning-based planes $\{\pi_{k,i}\}_{i \in \mathcal{I}_k^{\text{key}}}$, together with the corresponding parameterized sketch loops $\mathcal{K}_{k,i,j}$, the goal of the adaptive extrusion step is to select the appropriate slicing planes $\pi_{k,i}$ and estimate the associated extrusion parameters, namely the extrusion type, direction, and height $(t_{k,i,j}, \mathbf{v}_{k,i,j}, h_{k,i,j})$. The adaptive extrusion procedure first attempts to reconstruct the part using a single sketch-extrude operation by selecting one of the canonical planes $\{\pi_{k,r}\}_{r=1}^{3}$ and estimating the corresponding extrusion parameters. If the resulting solid does not yield a high-fidelity reconstruction with respect to the input part mesh $\mathbf{M}_k$, the learning-based multi-plane candidates are then considered, together with the estimation of their corresponding extrusion parameters.

\vspace{-0.1cm}
\paragraph{Single extrusion (canonical-based single plane).}

When a part can be represented by a single extrusion, candidate slicing planes are obtained from the canonical single-plane estimator 
$
\{\pi_{k,r}\}_{r=1}^{3}
=
\mathcal{E}_{\text{plane}}^{\text{single}}(\tilde{\mathbf{M}}_k),
$
where $\pi_{k,r} = (\mathbf{t}_k, \mathbf{R}_k \mathbf{a}_r)$ 
and $\mathbf{a}_r$ denotes the $r$-th canonical axis. For each $r \in \{1,2,3\}$, a single sketch extrude candidate $\mathbf{C}_k^{(r)}$
is constructed using the corresponding parametric sketch and extrusion direction $\mathbf{v}_{k}^{(r)} = \mathbf{R}_k \mathbf{a}_r$.
The extrusion height $h_{k}^{(r)}$ is initialized from the OBB extent along $\mathbf{a}_r$ and the type $t_{k}^{(r)} \in \{\textit{new},\textit{cut}\}$ is determined from 
loop nesting relationships as in~\cite{karadenizMiCADangelo}. The candidate is converted to a triangle mesh using a tessellation operator 
$
\hat{\mathbf{M}}_{k}^{(r)} = \mathcal{T}(\mathbf{C}_k^{(r)}).
$
The optimal axis is selected as
\begin{equation}
r^\star
=
\arg\min_{r \in \{1,2,3\}}
\texttt{CD}(\hat{\mathbf{M}}_{k}^{(r)}, \mathbf{M}_k) \ ,
\end{equation}
where $\texttt{CD}(\cdot,\cdot)$ denotes the Chamfer distance. If 
$
\texttt{CD}(\hat{\mathbf{M}}_{k}^{(r^\star)}, \mathbf{M}_k)
\leq \delta,
$
where $\delta$ is a fixed threshold set to $0.2$, the candidate is accepted and $\mathbf{C}_k = \mathbf{C}_k^{(r^\star)}$.
Otherwise, reconstruction proceeds to the multi-extrusion stage.

\vspace{-0.1cm}
\paragraph{Multi-extrusion (learning-based multi-planes).}

When single extrusion is insufficient to reconstruct the part, slicing planes are obtained via a learning-based plane detection $
\{\pi_{k,i}\}_{i\in\mathcal{I}_k^{\text{key}}}
=
\mathcal{E}_{\text{plane}}^{\text{multi}}(\tilde{\mathbf{M}}_k).
$ As in~\cite{karadenizMiCADangelo}, for each sketch $\mathcal{K}_{k,i,j}$, the corresponding extrusion 
parameters are defined, where $\mathbf{v}_{k,i,j}$ equals the normal of $\pi_{k,i}$,
and $t_{k,i,j} \in \{\textit{new},\textit{cut}\}$ is determined from 
loop nesting relationships. Extrusion heights $\{h_{k,i,j}\}$ are optimized via gradient descent similar to ~\cite{karadenizMiCADangelo}.

\vspace{-0.3cm}
\subsection{Part Assembly}

Given the reconstructed part-level CAD sequence
$
\mathbf{C}_k = \{(\mathcal{K}_{k,i,j}, \mathbf{e}_{k,i,j}) \mid i=1,\ldots,n_{\text{key}},\; j=1,\ldots,n_l^{k,i}\},
$
where $n_{\text{key}}$ and $n_l^{k,i}$ denote the number of key slicing planes and the corresponding number of loops, respectively, the complete CAD model is obtained by assembling all reconstructed parts in their original spatial configuration. The part-level sequences are merged to form the final CAD model.
\begin{equation}    
\mathbf{C}
=
\bigcup_{k=1}^{K}
\mathbf{C}_k \ .
\end{equation}

\section{\texttt{CC3D Cross-Section} Dataset}
\label{sec:cc3D}

To bridge the domain gap between clean CAD sketch images and 
scan-derived cross-section images, we construct a paired cross-section dataset 
from the original CC3D dataset~\cite{cherenkova2020pvdeconv,mallis2023sharp}. 
Given aligned CAD BRep (STEP) models and their corresponding scanned meshes, 
both representations are sliced using uniformly sampled planes along the 
canonical $x$-, $y$-, and $z$-axes. For each slicing plane, the BRep model 
yields parametric curves that are projected to 2D and converted into 
structured constrained sketches following the SketchGraphs format~\cite{seff2020sketchgraphs}. 
The corresponding mesh produces closed-loop curves, which are projected to 2D 
and rasterized into binary images. Both representations are centered and normalized within a unit bounding box, 
resulting in paired samples $(\mathbf{X}, \mathcal{K})$, where 
$\mathbf{X} \in \{0,1\}^{H \times W}$ denotes a scan-derived cross-section 
image and $\mathcal{K}$ the corresponding BRep-derived parametric sketch. 
The resulting dataset contains $44\mathrm{k}$ training and $2\mathrm{k}$ 
testing image–sketch pairs. The introduced 
\texttt{CC3D Cross-Section} dataset is used to fine-tune the sketch 
parameterization model and improve its robustness to scanning artifacts. 
Additional details about the dataset are provided in the supplementary material.

\section{Experiments}
\label{sec:experiments}

In this section, we describe the experimental setup and the experiments conducted to assess the performance of~\modelname.

\noindent\textbf{Evaluation Datasets.} We evaluate our method on the real-world CC3D~\cite{mallis2023sharp} testing set containing $2978$ samples and a subset of the Fusion 360 Assembly dataset~\cite{willis2021joinable} with $200$ assemblies containing $3780$ parts. The constrained sketch parameterization module is evaluated on the testing set of the introduced \texttt{CC3D Cross-Sections} dataset containing $2\mathrm{k}$ samples. Details of the evaluation datasets are provided in the supplementary.

\noindent\textbf{Metrics.} Median Chamfer Distance, median Edge Chamfer Distance (ECD), and Invalidity Ratio (IR) are used to evaluate the quality of 3D CAD reconstruction. Accuracy (Acc), Primitive F1 (PF1), Sketch Chamfer Distance (SCD), and number of connected components (NCC) are used for sketch-level evaluations. Segmentation IoU (Seg. IoU) is used for segmentation performance. Further details on the metrics are provided in the supplementary material.

\noindent\textbf{Implementation Details.} Input meshes are normalized to a unit bounding box and segmented using P3-SAM. Cross-sections are processed by a sketch parameterization network pretrained on SketchGraphs and fine-tuned on CC3D Cross-Sections. Reconstructed parts are instantiated and assembled in FreeCAD. Further implementation details are provided in the supplementary material.

\begin{table}[h]
\centering
\scriptsize
\setlength{\tabcolsep}{2.2pt}
\renewcommand{\arraystretch}{0.95}
\begin{tabularx}{\linewidth}{
l
*{3}{>{\centering\arraybackslash}X}
@{\hspace{8pt}}
*{3}{>{\centering\arraybackslash}X}
c
}
\toprule
\multirow{2}{*}{Method} 
& \multicolumn{3}{c}{CC3D} 
& \multicolumn{3}{c}{CC3D-Complex}\\
\cmidrule(lr){2-4} \cmidrule(lr){5-7} %
& CD$\downarrow$ & IR$\downarrow$ & ECD$\downarrow$
& CD$\downarrow$ & IR$\downarrow$ & ECD$\downarrow$
& \#Params \\
\midrule

CAD-SIGNet~\cite{khan2024cad} 
& 2.90 & 4.4 & 8.68
& 5.39 & 5.03 & 15.36
& 6.1M \\

MiCADangelo~\cite{karadenizMiCADangelo}  
& 1.69 & 2.2 & 5.93
& 3.37 & 3.2 & 8.18
& 59M \\

CAD-Recode~\cite{rukhovich2025cad}  
& 0.56 & 8.6 & 3.41
& 0.93 & 10.0 & 4.66
& 1.5B \\

Cadrille~\cite{kolodiazhnyi2025cadrille}  
& \textbf{0.44} & \textbf{0.5} & 3.65
& 0.74 & \textbf{0.5} & 4.91
& 2B \\

Ours  
& 0.48 & 1.6 & \textbf{2.78}
& \textbf{0.71} & 0.7 & \textbf{3.57}
& 171M \\

\bottomrule
\end{tabularx}

\caption{Quantitative Results on CC3D and CC3D-Complex.}
\label{tab:cc3d_combined}
\end{table}

\vspace{-0.6cm}
\subsection{Comparison with state-of-the-art}

The proposed approach is evaluated against CAD-SIGNet~\cite{khan2024cad}, MiCADangelo~\cite{karadenizMiCADangelo}, CAD-Recode~\cite{rukhovich2025cad}, and Cadrille~\cite{kolodiazhnyi2025cadrille} on the CC3D~\cite{mallis2023sharp} and Fusion360 Assembly datasets~\cite{willis2021joinable}. 

\begin{figure}[t!]
\setlength{\belowcaptionskip}{-0.1cm}
    \centering
    \includegraphics[width=\linewidth]{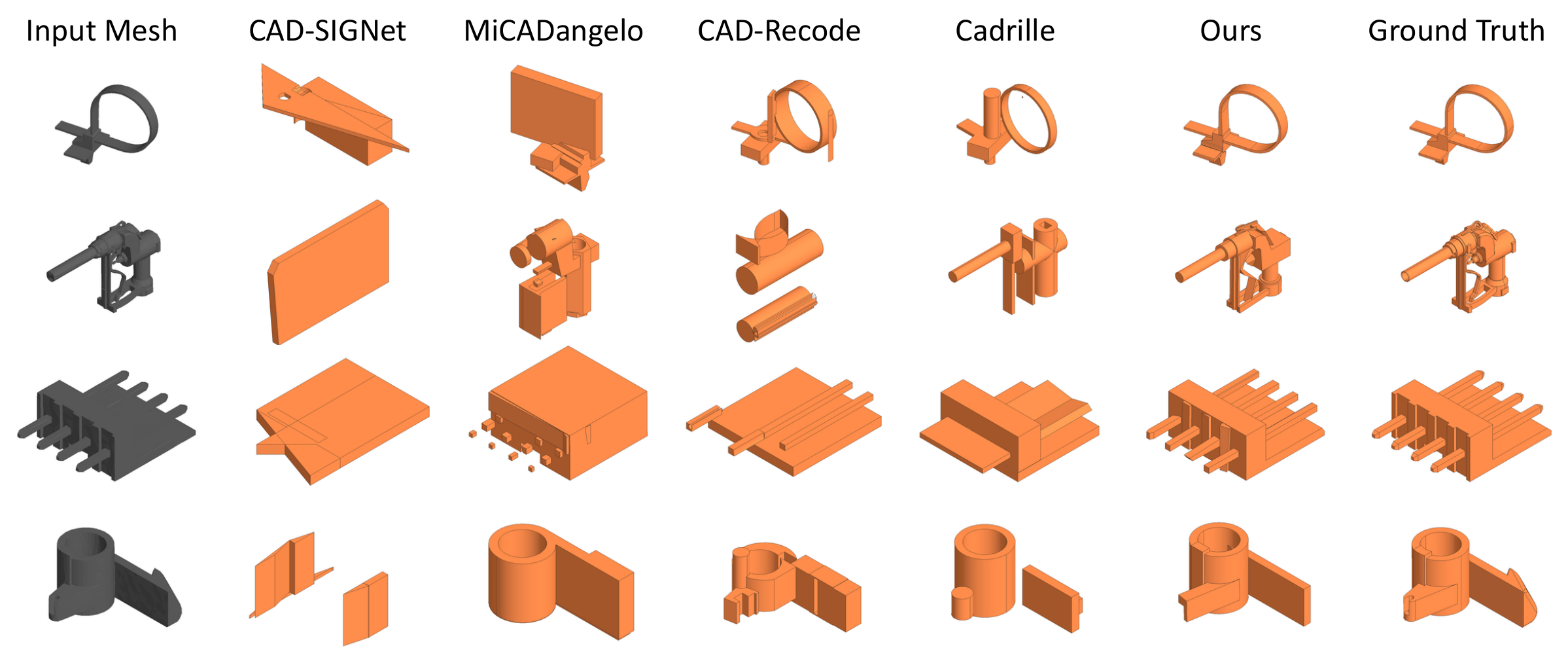}
    \caption{Qualitative comparison of our method with CAD-SIGNet~\cite{khan2024cad}, MiCADangelo~\cite{karadenizMiCADangelo}, CAD-Recode~\cite{rukhovich2025cad}, and Cadrille~\cite{kolodiazhnyi2025cadrille} on CC3D dataset.}
    \vspace{-0.1cm}
    \label{fig:cc3d_qual}
\end{figure}

\vspace{0.1cm}
\noindent We start with the evaluation on the CC3D dataset where the comparison is performed both on the full testing set ($2978$ samples) and a complex subset of the testing set (CC3D-Complex), where the number of operations to construct CAD models is more than $4$, resulting in $1528$ samples. Table~\ref{tab:cc3d_combined} shows the quantitative comparison with the state-of-the-art. On the full CC3D testing set, our method achieves the best Edge Chamfer Distance (ECD), indicating improved recovery of geometric structure, while maintaining competitive Chamfer Distance (CD) and Invalidity Ratio (IR) compared to a recent concurrent LLM-based approach~\cite{kolodiazhnyi2025cadrille}. Notably, this performance is achieved using only $171\mathrm{M}$ parameters, compared to $2\mathrm{B}$ parameters for Cadrille~\cite{kolodiazhnyi2025cadrille}, demonstrating substantially higher parameter efficiency. Furthermore, our method significantly outperforms the recent cross-section–based approach MiCADangelo~\cite{karadenizMiCADangelo}, highlighting the advantage of our part-decomposition-based reconstruction.  On the complex subset, the proposed method achieves the best CD and ECD. Figure~\ref{fig:cc3d_qual} provides a qualitative comparison of our method with other approaches on challenging samples. The proposed method yields high-fidelity CAD reconstructions that closely approximate the ground truth models.

\begin{table}[t]
\centering
\scriptsize
\setlength{\tabcolsep}{2.5pt}
\renewcommand{\arraystretch}{0.95}

\begin{tabularx}{\linewidth}{
@{}
l
*{3}{>{\centering\arraybackslash}X}
@{\hspace{7pt}}
*{3}{>{\centering\arraybackslash}X}
@{}
}
\toprule
\multirow{2}{*}{Method}
& \multicolumn{3}{c}{F360}
& \multicolumn{3}{c}{F360-Complex} \\
\cmidrule(lr){2-4}
\cmidrule(lr){5-7}
& CD$\downarrow$ & IR$\downarrow$ & ECD$\downarrow$
& CD$\downarrow$ & IR$\downarrow$ & ECD$\downarrow$ \\
\midrule

CAD-SIGNet~\cite{khan2024cad}
& 19.32 & 26.5 & 49.89
& 19.67 & 29.3 & 67.78 \\

MiCADangelo~\cite{karadenizMiCADangelo}
& 12.07 & 9.0 & 23.23
& 14.89 & 11.1 & 33.34 \\

CAD-Recode~\cite{rukhovich2025cad}
& 2.52 & 15.5 & 8.15
& 2.94 & 17.4 & 8.68 \\

Cadrille~\cite{kolodiazhnyi2025cadrille}
& 1.81 & 0.0 & 8.39
& 2.14 & 0.0 & 11.44 \\

Ours
& \textbf{1.57} & \textbf{0.0} & \textbf{5.28}
& \textbf{1.87} & \textbf{0.0} & \textbf{5.09} \\
\bottomrule
\end{tabularx}

\caption{Quantitative results on Fusion 360 Assembly and its complex subset.}
\label{tab:assembly_combined}
\vspace{-0.5cm}
\end{table}

\begin{figure}[h!]
    \centering
    
    \includegraphics[width=\linewidth]{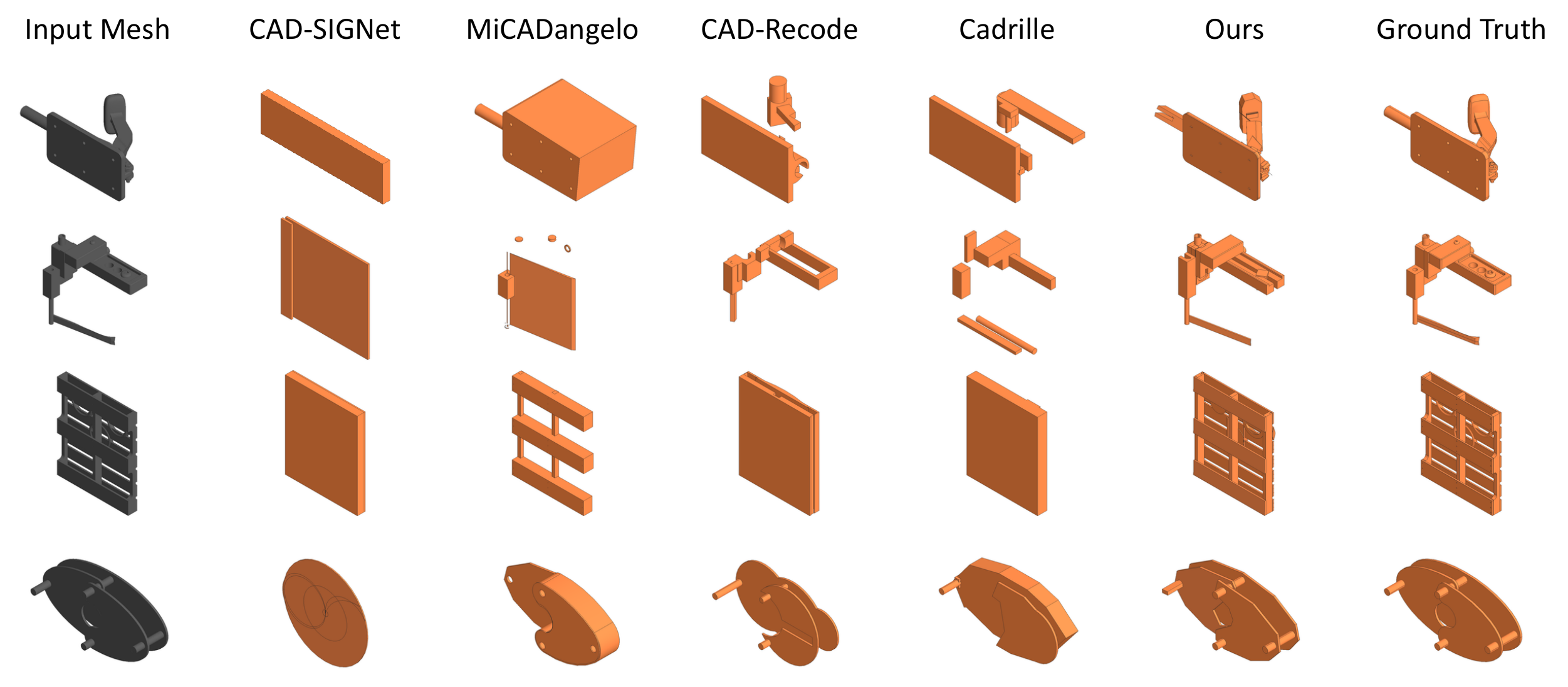}
    \caption{Qualitative comparison of our method with CAD-SIGNet~\cite{khan2024cad}, MiCADangelo~\cite{karadenizMiCADangelo}, CAD-Recode~\cite{rukhovich2025cad}, and Cadrille~\cite{kolodiazhnyi2025cadrille} on Fusion 360 Assembly dataset.}
    \label{fig:f360_assembly_qual}
    \vspace{-0.5cm}
\end{figure}

\noindent To evaluate performance on multi-part objects, we compare against prior work on the Fusion360 Assembly dataset~\cite{willis2021joinable}. Table~\ref{tab:assembly_combined} reports quantitative results on the selected subset ($200$ samples) and on a more complex subset of assemblies containing more than $4$ parts ($126$ samples). Our method achieves the best performance across all metrics in both settings, indicating stronger generalization to multi-part CAD models. Figure~\ref{fig:f360_assembly_qual} provides qualitative comparisons, where our reconstructions better preserve part geometry and overall assembly structure.

\noindent To assess robustness to out-of-distribution shapes containing extrusions non-aligned with canonical axes, we evaluate our method on a synthetic benchmark where sketches are randomly generated from primitive profiles (circles, polygons, and rectangles) and extruded along random 3D axes, producing stacks of primitives with varying orientations. Each sketch is instantiated on its corresponding oriented plane, extruded, and exported to form $200$ evaluation solids. As shown in Table~\ref{tab:directed_extrusions}, our method generalizes well to these extrusions, whereas competing approaches degrade substantially, reflecting their bias toward canonical-axis–aligned extrusion directions in the training data.

\vspace{-0.1cm}
\begin{table}[h!]
    \centering

    \scriptsize
    \small
    \begin{tabular}{lccc}
        \toprule
        Method & CD$\downarrow$ & ECD$\downarrow$ & IR$\downarrow$ \\
        \midrule
        CAD-SIGNet~\cite{khan2024cad}
        & 13.82 & 49.5 & 29.5 \\
        
        MiCADangelo~\cite{karadenizMiCADangelo}
        & 10.5 & 36.16 & 2.5 \\
        
        CAD-Recode~\cite{rukhovich2025cad}
        & 2.1 & 4.87 & 12.0 \\
        
        Cadrille~\cite{kolodiazhnyi2025cadrille}
        & 1.37 & 4.31 & 0.0 \\
        
        Ours
        & \textbf{0.08} & \textbf{0.24} & \textbf{0.0} \\
        \bottomrule
    \end{tabular}
    \vspace{-0.1cm}
    \caption{Quantitative results on extrusions non-aligned with canonical axes.}
    \label{tab:directed_extrusions}

\end{table}

\vspace{-0.7cm}
\subsection{Ablation Study}

\noindent \textbf{Cross-Section Parameterization.} Table~\ref{tab:sketch_combined} reports the performance of our CAD sketch parameterization model on the testing set of the introduced \texttt{CC3D Cross-Sections} dataset. Compared to training only on SketchGraphs (SG)~\cite{karadeniz2024davinci} or using synthetic noisy augmentations (SG+Synt)~\cite{karadenizMiCADangelo}, our model, fine-tuned on the proposed \texttt{CC3D Cross-Sections} dataset (SG + CC3D-CS), achieves the best performance on the cross-section parameterization task. Notably, improvements in sketch parameterization are also reflected in the 3D reconstruction performance on testing set of CC3D~\cite{mallis2023sharp}.

\begin{table}[h]
\centering
\scriptsize
\setlength{\tabcolsep}{2.2pt}

\renewcommand{\arraystretch}{0.95}

\begin{tabular}{lccccccc}
\toprule
& \multicolumn{4}{c}{CAD Sketch Parameterization} & \multicolumn{3}{c}{3D Reconstruction} \\
\cmidrule(lr){2-5} \cmidrule(lr){6-8}
Training Dataset & Acc$\uparrow$ & PF1$\uparrow$ & SCD$\downarrow$ & NCC$\downarrow$ & CD$\downarrow$ & IR$\downarrow$ & ECD$\downarrow$ \\
\midrule
SG~\cite{seff2020sketchgraphs}
& 0.851 & 0.427 & 2.24 & 1.95
& 0.59 & 2.8 & 3.17 \\

SG + Synt~\cite{karadenizMiCADangelo}
& 0.880 & 0.550 & 0.62 & 1.31
& 0.52 & \textbf{1.3} & 3.01 \\

SG + CC3D-CS (Ours)
& \textbf{0.928} & \textbf{0.844} & \textbf{0.251} & \textbf{1.10}
& \textbf{0.48} & 1.6 & \textbf{2.78} \\
\bottomrule
\end{tabular}
\vspace{-0.1cm}
\caption{Effect of different training datasets on sketch parameterization and their impact on 3D reconstruction performance on CC3D.}
\vspace{-0.3cm}
\label{tab:sketch_combined}
\end{table}

\vspace{-0.1cm}
\noindent \textbf{Impact of Part Segmentation Methods.} 
As shown in Table~\ref{tab:segmentation_extrusion} \textbf{(top)}, we evaluate several 3D part segmentation methods for CAD reconstruction on the Fusion360 Assembly dataset~\cite{willis2021joinable}. The compared methods include PointNeXt~\cite{qian2022pointnext}, trained on the 3DCoMPaT++~\cite{li2022_3dcompat,slim2023_3dcompatplus} dataset of CAD models; the 2D-SAM–based approach SAMPart3D~\cite{yang2024sampart3d}; and recent native 3D foundation models, namely PartSAM~\cite{zhu2025partsam} and P3-SAM~\cite{ma2025p3}. We observe that native 3D foundational segmentation models, PartSAM~\cite{zhu2025partsam} and P3-SAM~\cite{ma2025p3} (achieving the better Seg. IoU results), yield better 3D reconstruction performance than segmentation models derived from 2D foundational models (SAMPart3D~\cite{yang2024sampart3d}). In contrast, the dedicated segmentation model PointNeXt~\cite{qian2022pointnext}, trained for part segmentation on a CAD dataset~\cite{li2022_3dcompat}, underperforms compared to these foundation-based approaches.

\noindent \textbf{Multi-Extrusion vs. Single Extrusion.} Table~\ref{tab:segmentation_extrusion} \textbf{(bottom)} compares the adaptive extrusion approach with a single-extrusion baseline on the CC3D dataset~\cite{mallis2023sharp}. The baseline reconstructs each part using only the canonical planes. In contrast, the full method additionally enables learning-based plane detection and a multi-extrusion optimization stage. Allowing multiple sketch–extrude operations per part improves reconstruction performance.

\begin{table}[t]
    \centering
    \small
    \setlength{\tabcolsep}{4pt}

    \begin{tabular}{lcccc}
        \toprule
        Method & Seg. IoU$\uparrow$ & CD$\downarrow$ & IR$\downarrow$ & ECD$\downarrow$ \\
        \midrule
    
        \multicolumn{4}{l}{\textit{Part segmentation on Fusion360 Assembly}} \\
    
        PointNeXt~\cite{qian2022pointnext}
        & 43.1 & 3.61 & 10.0 & 9.54 \\
    
        SAMPart3D~\cite{yang2024sampart3d}
        & 52.5 & 2.90 & 6.5 & 9.47 \\
    
        PartSAM~\cite{zhu2025partsam}
        & 72.5 & 2.08 & 6.0 & 5.51 \\
    
        P3-SAM~\cite{ma2025p3}
        & \textbf{77.6} & \textbf{1.57} & \textbf{0.0} & \textbf{5.28} \\
    
        \midrule
        \multicolumn{5}{l}{\textit{Impact of Extrusion strategy on CC3D}} \\
    
        Single-extrusion
        & -- & 0.51 & \textbf{1.5} & 3.00 \\
    
        Multi-extrusion
        & -- & \textbf{0.48} & 1.6 & \textbf{2.78} \\
    
        \bottomrule
    \end{tabular}
    
    \caption{Ablation studies on part segmentation and extrusion strategy.}

    \vspace{-0.7cm}

    \label{tab:segmentation_extrusion}
\end{table}

\vspace{-0.3cm}
\section{Conclusion}
\label{sec:conclusion}

In this work, we presented~\modelname, a part-aware CAD reverse engineering method that reconstructs parametric CAD models from 3D scans by decomposing complex objects into structurally meaningful components. By combining foundation-scale 3D part segmentation with a cross-section-driven reconstruction pipeline, our method enables robust sketch parameterization and extrusion at the part level. We further introduced the \texttt{CC3D Cross-Sections} dataset, which bridges the domain gap between clean CAD sketches and scan-derived cross-sections, improving generalization to realistic scanning artifacts. Extensive experiments demonstrate that the proposed approach achieves state-of-the-art performance on the CC3D-Complex and Fusion360-Assembly benchmarks while remaining scalable to complex multi-part objects. Similar to prior extrusion-based methods~\cite{khan2024cad, karadenizMiCADangelo, rukhovich2025cad, kolodiazhnyi2025cadrille}, the current framework supports only extrusion operations; revolutions and sweeps are not modeled. In addition, parts are assembled via direct fusion rather than explicit joint relationships~\cite{willis2021joinable}. Future work will extend the framework to support additional operations and assembly joint prediction.

\section{Acknowledgements}
This work is supported by the National Research Fund (FNR), Luxembourg, under the BRIDGES2021/IS/16849599/FREE-3D project and by Artec3D.

\appendix

\bibliography{main}
\clearpage
\section{Supplementary Material}

This supplementary document provides additional technical details and expanded qualitative evaluations for \modelname{}.

\subsection{Datasets}
\label{sec:evaluation_datasets}

This section describes the evaluation datasets and provides details of the introduced \texttt{CC3D Cross-Section} dataset.

\paragraph{CC3D Test Set.}
We evaluate our method on the CC3D~\cite{mallis2023sharp} dataset, which consists of paired real-world scans represented as meshes and their corresponding CAD models. The CC3D test set contains $2978$ samples. We further evaluate our method on a more complex subset (CC3D-Complex), where the number of construction steps exceeds $4$, containing $1528$ samples.

\paragraph{Fusion 360 Assembly Test Set.}
To evaluate our approach on assemblies, we use a subset of the Fusion 360 Assembly dataset~\cite{willis2021joinable}. This subset contains $200$ assemblies comprising a total of $3780$ parts. We also evaluate on a more complex subset of $126$ assemblies, each containing more than $4$ parts. These assemblies represent realistic mechanical designs composed of multiple components.

\paragraph{Non-Axis-Aligned Extrusions Test Set.}
To evaluate robustness to extrusions that are not aligned with canonical axes, we construct a synthetic benchmark of procedurally generated CAD models. Each sample is created by stacking a sequence of primitive extrusions with randomly oriented directions in 3D. For each extrusion step, a sketch profile is randomly sampled from primitive shapes including circles, regular polygons ($3$--$9$ sides), and rectangles with randomized sizes. The sketch is instantiated on a plane whose normal corresponds to the sampled extrusion direction and is extruded by a length sampled from a uniform distribution in the range $[0.2, 1.2]$. Successive sketches are placed on the top face of the previous extrusion, forming stacks of primitives with varying orientations and geometric complexity. The extrusion directions are sampled to allow tilts of up to $35^\circ$ from the previous direction, producing non-axis-aligned geometries that differ from the canonical orientations common in many CAD datasets. Using this procedure, we generate $200$ solids, each consisting of $2$--$7$ sequential extrusion operations with varying profiles, scales, and extrusion lengths, which are exported as STEP and mesh files and used as an out-of-distribution benchmark.

\begin{figure}[h!]
    \centering
    \includegraphics[width=\linewidth]{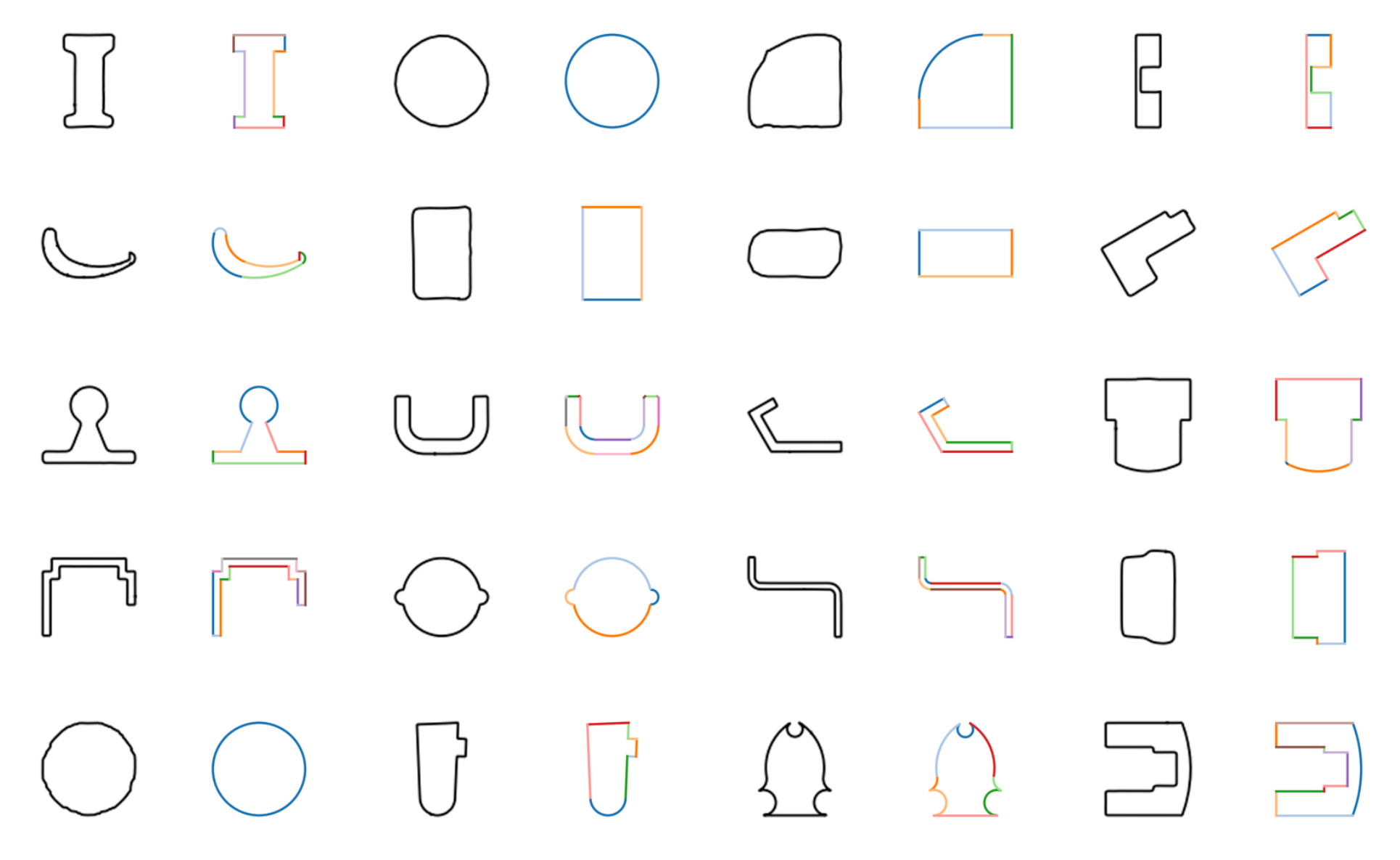}
    \caption{Example cross-section image and parametric sketch pairs from the \texttt{CC3D Cross-Section} dataset.}
    \label{fig:cc3d_sketches_supp}
\end{figure}

\paragraph{CC3D Cross-Section Dataset.}
We introduce the \texttt{CC3D Cross-Section} dataset, constructed from aligned CAD BRep models and real scanned meshes provided in CC3D~\cite{cherenkova2020pvdeconv,mallis2023sharp}. For each aligned BRep--scan pair, we sample $3$ uniformly spaced slicing planes along each of the canonical \(x\)-, \(y\)-, and \(z\)-axes and intersect both representations with these planes. From the BRep representation, each valid connected cross-sectional contour is converted into a parametric sketch of supported primitives in the SketchGraphs format~\cite{seff2020sketchgraphs}, while the corresponding mesh cross-section is projected onto the slicing plane and rasterized into a binary image. We filter samples by restricting the number of sketch entities to at most $16$. To ensure geometric consistency between the scan-derived cross-section images and the parametric sketches, we compute the sketch Chamfer distance (SCD) between their rasterized representations and discard pairs with $\mathrm{SCD} > 1$. Duplicate sketches are removed. The resulting dataset consists of paired samples comprising a scan-derived cross-section image and its corresponding parametric sketch, which are used to fine-tune the sketch parameterization model and improve robustness to scanning artifacts. Example sketches from the dataset are shown in Figure~\ref{fig:cc3d_sketches_supp}. The final dataset contains $44\mathrm{k}$ training pairs and $2\mathrm{k}$ test pairs, with an average of $8.3$ primitives per sketch.

\vspace{-0.5cm}

\vspace{-0.5cm}
\subsection{Evaluation Metrics}
\label{sec:evaluation_metrics}

Reconstruction quality is evaluated using both geometric metrics in 3D and sketch-level metrics.

\paragraph{Chamfer Distance (CD).}
To evaluate the geometric fidelity of the reconstructed shapes, we compute the Chamfer Distance between the predicted and ground-truth meshes obtained after converting the STEP files. We uniformly sample $N=8192$ surface points from predicted and ground truth meshes, producing two point sets
\[
\mathcal{P} = \{\mathbf{x}_i\}_{i=1}^{N}, \qquad 
\mathcal{Q} = \{\mathbf{y}_i\}_{i=1}^{N},
\]
with $\mathbf{x}_i, \mathbf{y}_i \in \mathbb{R}^3$. The sampled point clouds are centered and normalized to unit scale. The Chamfer Distance is then computed as
\begin{equation}
\operatorname{CD}(\mathcal{P},\mathcal{Q}) =
\frac{1}{2N}
\sum_{i=1}^{N}\min_{\mathbf{y}_j \in \mathcal{Q}}
\|\mathbf{x}_i - \mathbf{y}_j\|_2^2
+
\frac{1}{2N}
\sum_{j=1}^{N}\min_{\mathbf{x}_i \in \mathcal{P}}
\|\mathbf{y}_j - \mathbf{x}_i\|_2^2 .
\end{equation}
We report the median CD across the evaluation set to reduce the influence of outlier reconstructions.

\paragraph{Edge Chamfer Distance (ECD).}
While CD evaluates overall surface geometry, Edge Chamfer Distance focuses specifically on boundary curves and sharp features. For each predicted and ground-truth CAD model represented as \texttt{STEP} files, we sample $M=4096$ points uniformly along all edge curves. This yields two edge point sets
\[
\mathcal{E}_p = \{\mathbf{e}_i\}_{i=1}^{M}, \qquad
\mathcal{E}_g = \{\mathbf{e}'_i\}_{i=1}^{M},
\]
where $\mathbf{e}_i, \mathbf{e}'_i \in \mathbb{R}^3$. After centering and normalizing the point clouds, the bidirectional Chamfer distance between the edge point sets is computed analogously to CD. The median ECD over all evaluation samples is reported.

\paragraph{Invalidity Ratio (IR).}
The Invalidity Ratio measures the proportion of predicted CAD models that cannot be successfully converted into valid mesh representations.

\paragraph{Accuracy (Acc).}
Accuracy measures the correctness of predicted primitive parameters at the token level. 
Since the ordering of predicted primitives may differ from the ground truth, an assignment 
between predicted and ground-truth primitives is first recovered using bipartite matching. 
This produces a permutation $\sigma \in S_{n_z}$ that aligns predicted primitives with the 
corresponding ground-truth primitives. Let $t_i^j$ denote the $j$-th token of the 
$i$-th ground-truth primitive and $\hat{t}_{\sigma(i)}^j$ the corresponding token of the 
matched predicted primitive. Accuracy is then defined as
\begin{equation}
\operatorname{Acc} =
\frac{1}{T_{\text{Acc}}}
\sum_{i=1}^{n_z}
\sum_{j=1}^{8}
\mathbf{1}{1}[t_i^j > 0]\;
\mathbf{1}{1}[t_i^j = \hat{t}_{\sigma(i)}^j],
\end{equation}
where $n_z$ denotes the number of primitives in the $z$-th test sample and 
$\mathbf{1}{1}[\cdot]$ is the indicator function. The normalization term $T_{\text{Acc}}$ 
corresponds to the number of valid (non-padding) tokens in the ground-truth sequence.

\paragraph{Primitive F1 (PF1).}
Primitive F1 evaluates the correctness of predicted sketch primitives. 
A primitive is considered correct if its predicted type matches the ground-truth type and all parameters lie within $5$ quantization units of the reference values. Primitive parameters are discretized using 6-bit quantization, consistent with the scheme used by the sketch parameterization network. The PF1 score is computed as the F1 score over the set of predicted primitives.

\paragraph{Number of Connected Components (NCC).}
The Number of Connected Components measures the structural coherence of predicted sketches, where a component is defined as a set of primitives whose start and end points coincide. Endpoints that lie within a small tolerance ($0.01$) are considered identical to account for numerical noise. Primitive types without explicit endpoints (e.g., circles) are treated as individual components. Ideally, a predicted sketch forms a single connected structure, yielding an NCC value of $1$.

\paragraph{Sketch Chamfer Distance (SCD).}
To evaluate geometric alignment between predicted and reference sketches, we compute a 2D Chamfer distance between their rasterized representations.
Each sketch is rendered as a binary image of resolution $128 \times 128$ after centering and normalizing it to a unit bounding box. The coordinates of foreground pixels are extracted and normalized by the image resolution:
\[
\mathcal{S}_g = \{\mathbf{u}_i\}_{i=1}^{N_g}, \qquad
\mathcal{S}_p = \{\mathbf{v}_i\}_{i=1}^{N_p},
\]
where $\mathbf{u}_i, \mathbf{v}_i \in [0,1]^2$. The Sketch Chamfer Distance is defined as the Chamfer distance (CD) between these two point sets.

\subsection{Additional Implementation Details}
\label{sec:implementation_details}

Input meshes are centered and normalized to a unit bounding box. Part segmentation is performed using P3-SAM~\cite{ma2025p3} on \(30\mathrm{k}\) sampled surface points with a mask confidence threshold of \(0.95\), and part submeshes are extracted from the predicted face labels. Part meshes with volume below \(0.1\) are discarded. For plane detection, \(40\) evenly spaced cross-sectional slices are taken by the learned plane detection network after rotating each part into a canonical frame by computing its oriented bounding box. The resulting cross-sections corresponding to the key slice planes are rasterized at a resolution of \(128 \times 128\) and processed by the sketch parameterization network of~\cite{karadeniz2024davinci}, which is pretrained on SketchGraphs~\cite{seff2020sketchgraphs} and fine-tuned for \(20\) epochs with a learning rate of \(3\times10^{-5}\). The predicted sketches are postprocessed through a sequence of cleanup and refinement steps. Specifically, duplicate and degenerate primitives are removed, and collinear overlapping line segments are merged. The maximum number of primitives per parametric sketch is limited to \(16\). Extrusion parameters are optimized using differentiable extrusion over \(8\) candidate extrusion directions for \(200\) iterations with a learning rate of \(2\times10^{-3}\). Finally, reconstructed parts are instantiated and combined using Boolean operations in FreeCAD through its Python API~\cite{FreeCAD}.

\subsection{Additional Experiments}
\label{sec:additional_ablations}

\begin{figure}[h!]
    \centering
    \includegraphics[width=\linewidth]{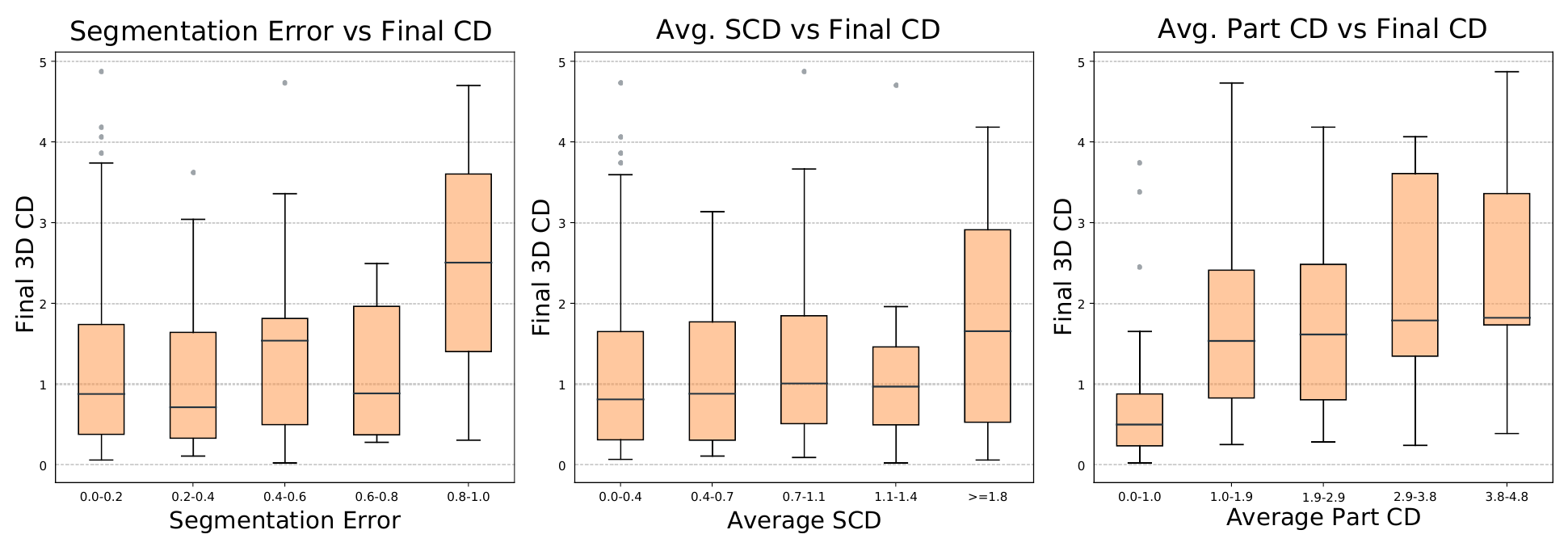}
    \caption{Relationship between intermediate pipeline errors and final reconstruction CD on Fusion 360 Assembly.}
    \label{fig:per_component_supp}
\end{figure}

\paragraph{Error Propagation Across Pipeline Stages.}
To quantify how intermediate inaccuracies affect the final reconstruction, we analyze three stage-level error indicators on Fusion 360 Assembly: (i) segmentation error, measured as \(1-\mathrm{IoU}\) against ground-truth parts; (ii) sketch parameterization error, measured using SCD between predicted sketches and rasterized cross-section sketches extracted from the input meshes; and (iii) average part-level CD between reconstructed parts and their corresponding input parts, which jointly captures errors from cross-section extraction, sketch parameterization, and extrusion fitting. Figure~\ref{fig:per_component_supp} shows that the final CD remains relatively stable under moderate segmentation and sketch errors, while part-level CD exhibits a stronger monotonic relationship with the final reconstruction error. These results indicate that the pipeline degrades gradually rather than compounding moderate intermediate inaccuracies catastrophically.

\paragraph{Cross-Dataset Generalization of Sketch Parameterization.}
We evaluate whether fine-tuning on \texttt{CC3D Cross-Sections} transfers to clean cross-dataset inputs using approximately (600) cross-section--sketch pairs extracted from Fusion 360 Assembly, which is used exclusively at test time. As shown in Table~\ref{tab:sketch_f360_supp}, the fine-tuned model achieves modest improvements in sketch-level SCD and PF1, together with small gains in downstream CD and ECD. These gains are smaller than those observed on CC3D because Fusion 360 Assembly meshes are triangulated from clean BRep models and do not contain the scanning artifacts specifically targeted by \texttt{CC3D Cross-Sections}. Nevertheless, the results show that training on scan-derived cross-sections preserves performance on clean inputs while providing a cross-dataset benefit.

\begin{table}[h!]
\centering
\scriptsize
\setlength{\tabcolsep}{2.4pt}
\caption{Effect of sketch training data on Fusion 360 Assembly.}
\label{tab:sketch_f360_supp}
\resizebox{\linewidth}{!}{%
\begin{tabular}{lccccccc}
\toprule
& \multicolumn{4}{c}{CAD Sketch Parameterization} & \multicolumn{3}{c}{3D Reconstruction} \\
\cmidrule(lr){2-5} \cmidrule(lr){6-8}
Training Dataset & Acc$\uparrow$ & PF1$\uparrow$ & SCD$\downarrow$ & NCC$\downarrow$ & CD$\downarrow$ & IR$\downarrow$ & ECD$\downarrow$ \\
\midrule
SG
& 0.907 & 0.697 & 0.746 & 2.32
& 1.64 & 1.0 & 6.06 \\
SG + Synt
& 0.933 & 0.764 & 0.262 & \textbf{1.39}
& 1.58 & \textbf{0.0} & 5.42 \\
SG + CC3D-CS
& \textbf{0.935} & \textbf{0.852} & \textbf{0.217} & 1.70
& \textbf{1.57} & \textbf{0.0} & \textbf{5.37} \\
\bottomrule
\end{tabular}}
\end{table}

\subsection{Additional Dataset and Runtime Clarifications}
\label{sec:additional_clarifications}

\paragraph{Input Mesh Provenance.}
CC3D contains real-world scanned meshes with acquisition artifacts, whereas the Fusion 360 Assembly inputs are meshes triangulated from BRep models. The same reconstruction pipeline is applied to both datasets without modification. Fusion 360 Assembly is not used for fine-tuning and therefore serves as a cross-dataset evaluation setting.

\paragraph{Constraint Annotations.}
The parametric sketches extracted for \texttt{CC3D Cross-Sections} contain coincident constraints. The sketch parameterization model retains its broader constraint prediction capability through pretraining on SketchGraphs and continued sampling of SketchGraphs during fine-tuning.

\paragraph{Part Filtering Statistics.}
Part meshes with volume below \(0.1\) are removed before reconstruction. Across the evaluation data, this filtering discards approximately \(0.17\%\) of the predicted parts.

\paragraph{Runtime.}
Runtime is measured on a system equipped with an NVIDIA RTX A4000 GPU and an Intel Xeon Gold 6226R CPU. Averaged over the evaluation set, part segmentation requires approximately 12 seconds per input, while the subsequent reconstruction stages require approximately 15 seconds, resulting in a total runtime of approximately 27 seconds per assembly.

\subsection{Failure Case Analysis}
\label{sec:limitations}

\begin{figure}[h!]
    \centering
    \includegraphics[width=\linewidth]{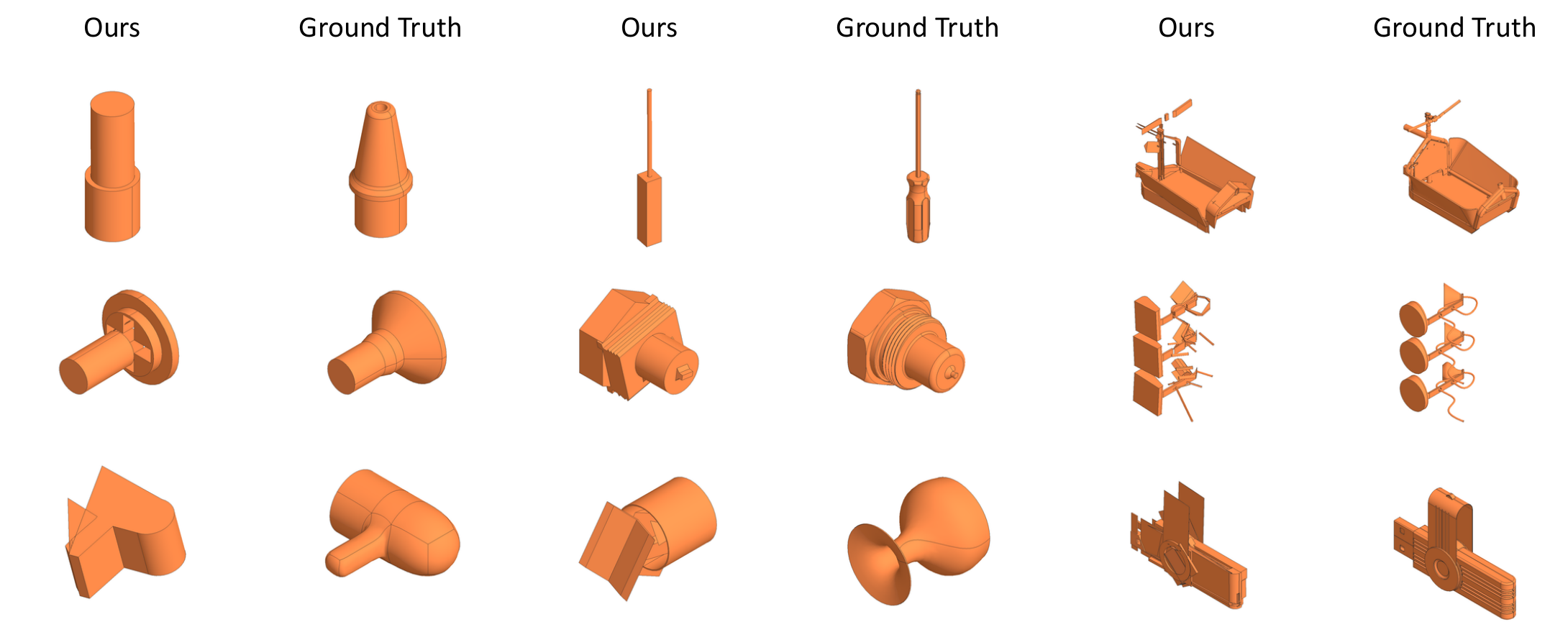}
    \caption{Representative failure cases of \modelname{}.}
    \label{fig:failure_cases}
\end{figure}

We analyze common failure cases of \modelname{}. 
Figure~\ref{fig:failure_cases} shows representative examples where the reconstruction pipeline produces suboptimal results. 
Some failures arise from unsupported CAD operations, such as revolutions, lofts, or sweeps, which cannot be represented by our extrusion-based formulation. 
Additional failures occur when the reconstructed assembly contains disconnected components, resulting in incomplete geometry. 
Finally, very thin structures such as chamfers or small fillets are often difficult to recover from scan data and may therefore be missed during reconstruction.

\subsection{Additional Qualitative Results}
\label{sec:more_qualitative}

\begin{figure*}[h!]
    \centering
    \includegraphics[width=0.9\linewidth]{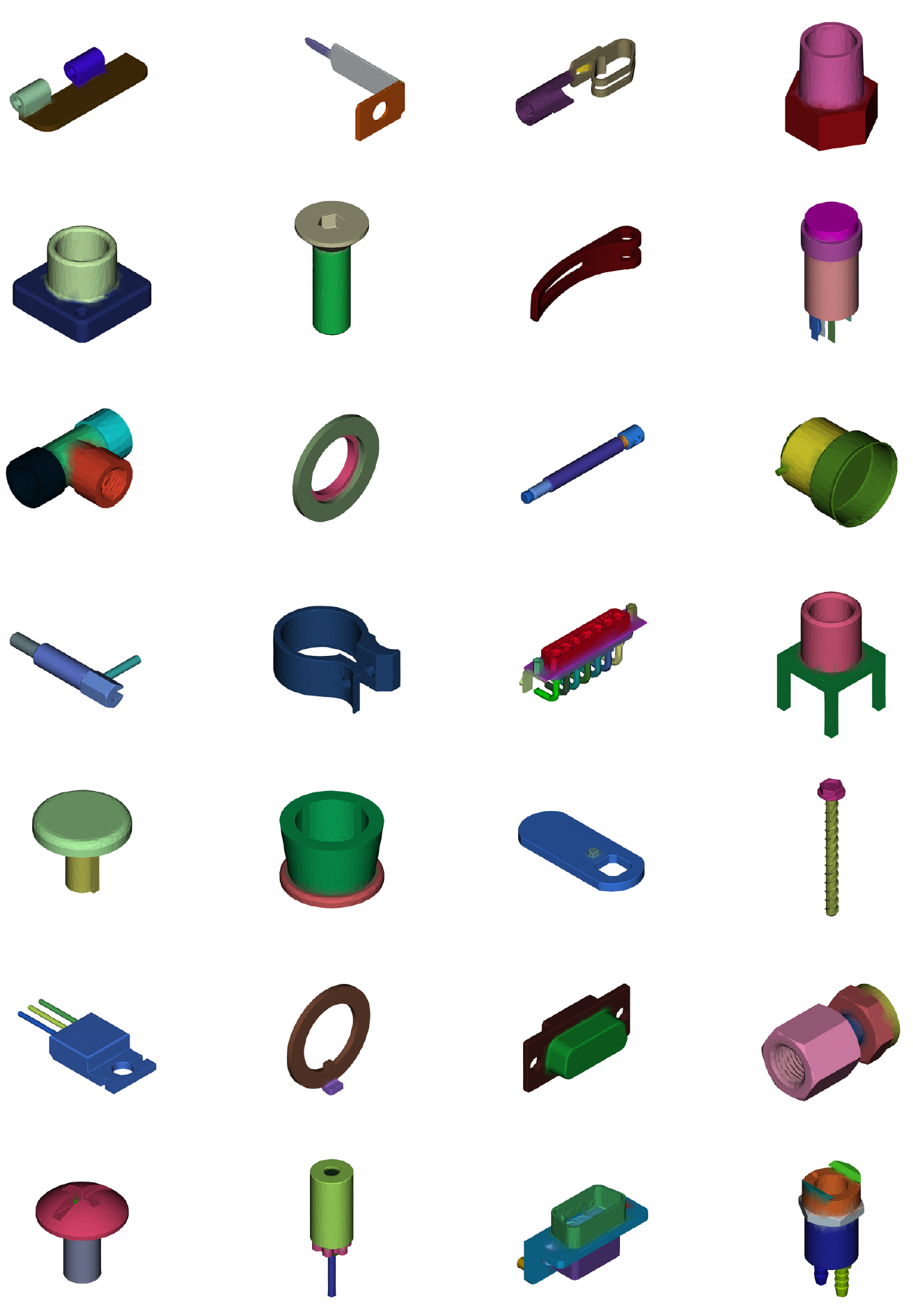}
    \caption{Example part segmentations from the CC3D dataset using P3-SAM~\cite{ma2025p3}.}
    \label{fig:cc3d_seg}
\end{figure*}

\begin{figure*}[h!]
    \centering
    \includegraphics[width=0.85\linewidth]{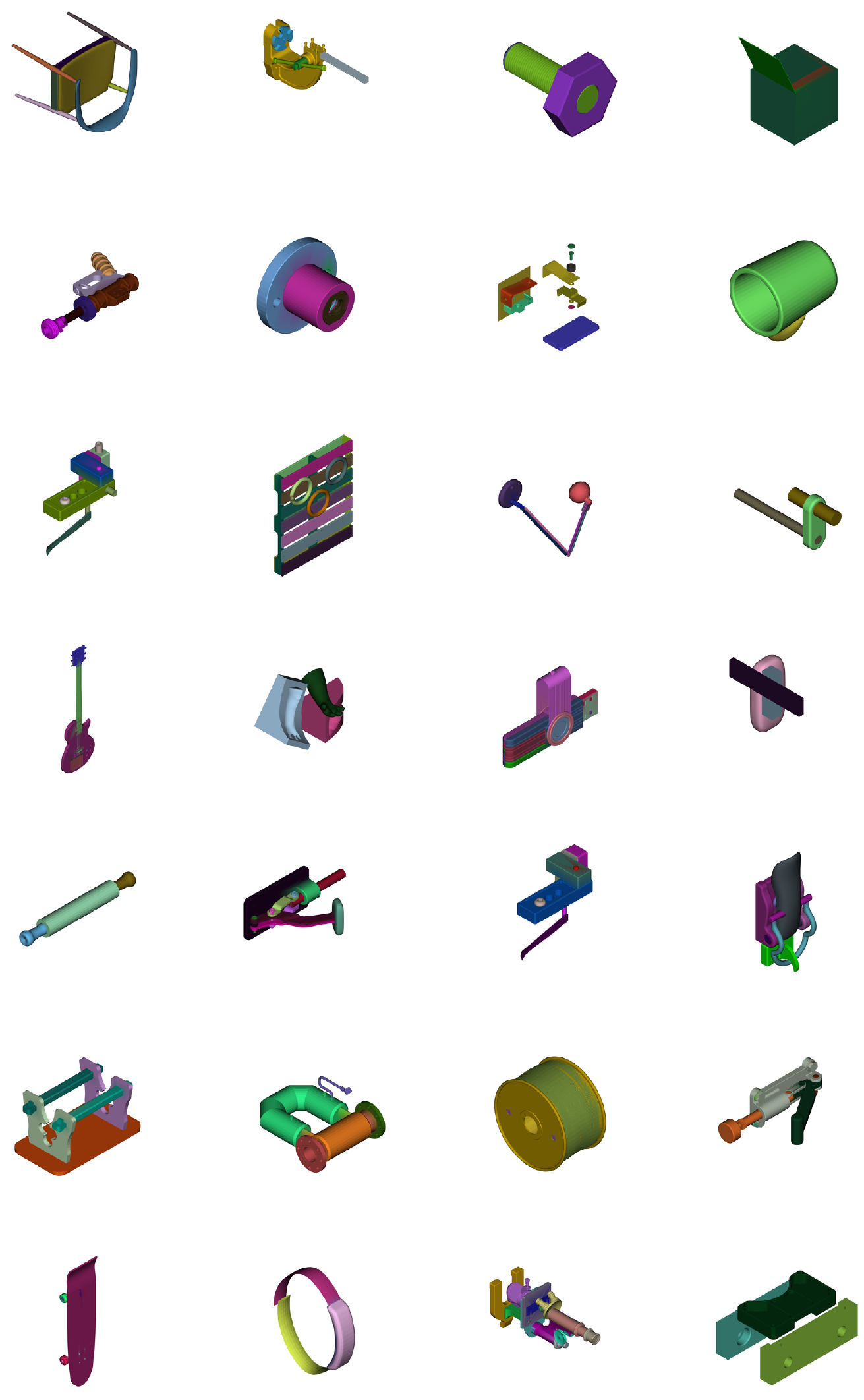}
    \caption{Example part segmentations from the F360 Assembly dataset using P3-SAM~\cite{ma2025p3}.}
    \label{fig:f360_seg}
\end{figure*}

\begin{figure*}[p]
    \centering
    \vfill
    \includegraphics[width=\linewidth]{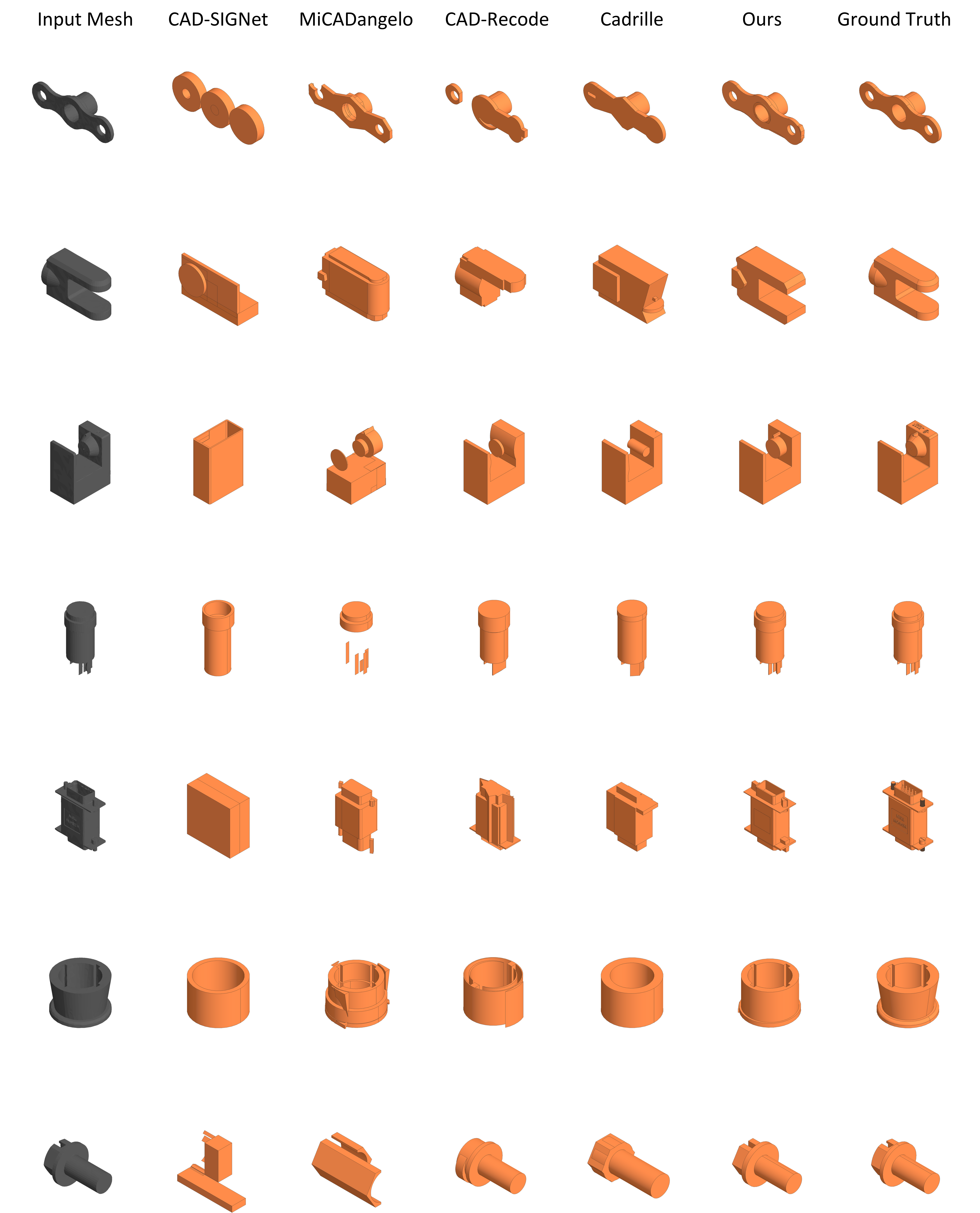}
    \caption{More qualitative comparisons with CAD-SIGNet~\cite{khan2024cad}, MiCADangelo~\cite{karadenizMiCADangelo}, CAD-Recode~\cite{rukhovich2025cad}, and Cadrille~\cite{kolodiazhnyi2025cadrille} on CC3D dataset.}
    \label{fig:cc3d_qual_more}
    \vfill
\end{figure*}

\begin{figure*}[p]
    \centering
    \vfill
    \includegraphics[width=\linewidth]{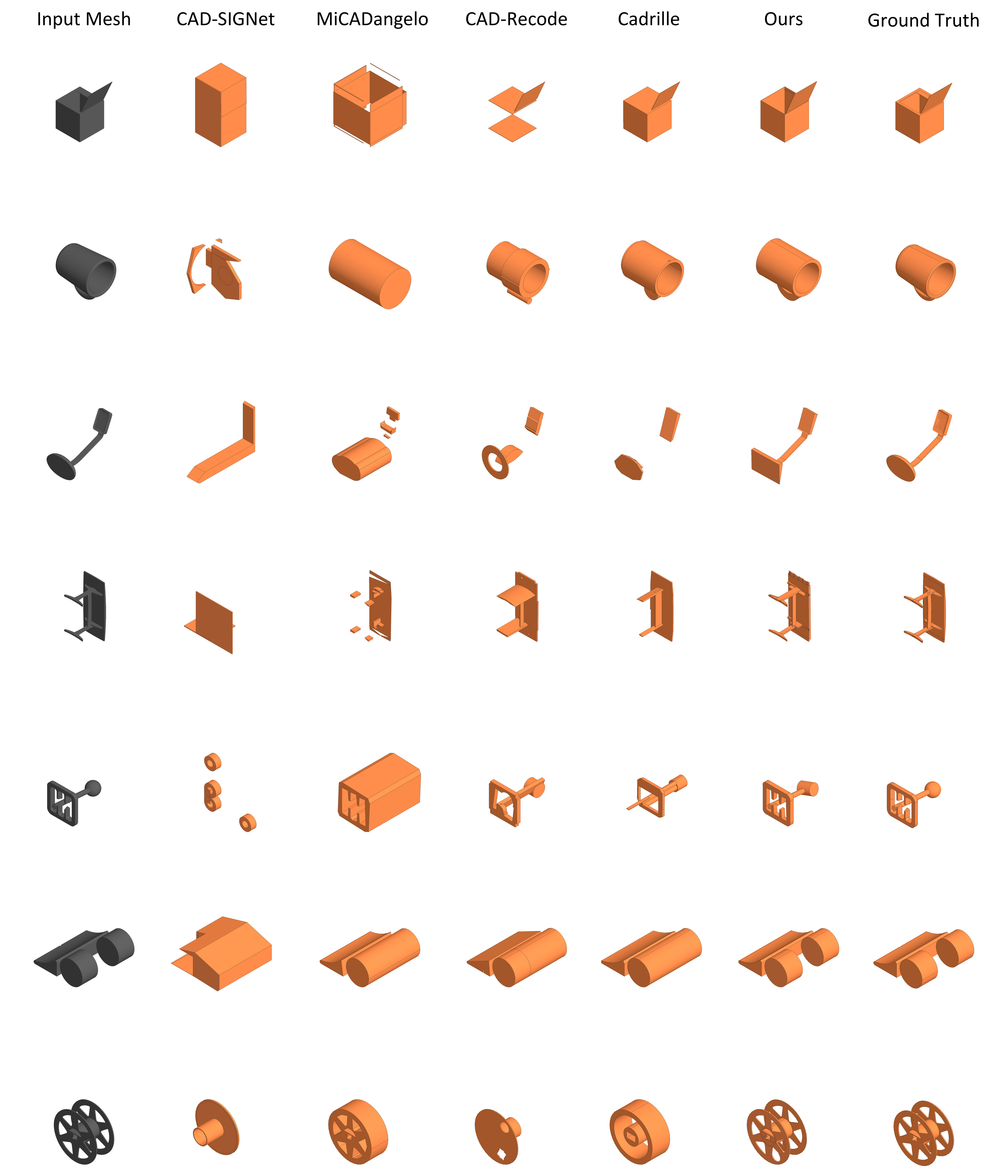}
    \caption{More qualitative comparisons with CAD-SIGNet~\cite{khan2024cad}, MiCADangelo~\cite{karadenizMiCADangelo}, CAD-Recode~\cite{rukhovich2025cad}, and Cadrille~\cite{kolodiazhnyi2025cadrille} on Fusion 360 Assembly dataset.}
    \label{fig:f360_qual_more}
    \vfill
\end{figure*}

\begin{figure*}[h!]
    \centering
    \includegraphics[width=0.9\linewidth]{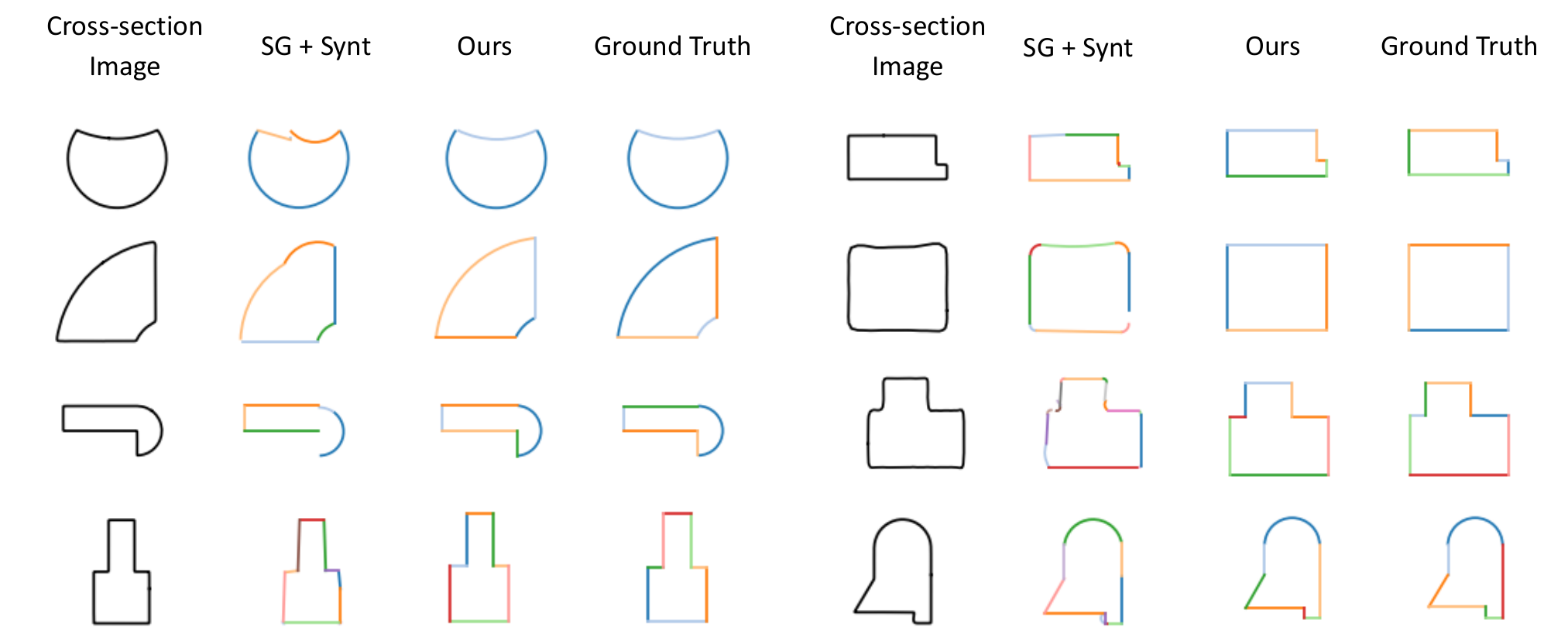}
    \caption{Qualitative comparison of sketch parameterization between our model (SketchGraphs pre-training + CC3D Cross-Section fine-tuning) and~\cite{karadenizMiCADangelo}, which is trained on SketchGraphs with synthetic noise (SG+Synt).}
    \label{fig:cs_qual}
\end{figure*}

This section provides further qualitative results along with visualizations. Figure~\ref{fig:cs_qual} shows qualitative results of the sketch parameterization model. 
Figure~\ref{fig:cc3d_seg} and Figure~\ref{fig:f360_seg} present example part segmentations from the CC3D and Fusion 360 Assembly datasets. 
Figure~\ref{fig:cc3d_qual_more} and Figure~\ref{fig:f360_qual_more} provide further qualitative comparisons on the CC3D and Fusion 360 Assembly datasets.

\vspace{-0.5cm}

\end{document}